%% file: paper.tex
\documentclass[10pt,twocolumn,letterpaper]{article}

\usepackage{cvpr}              


\definecolor{cvprblue}{rgb}{0.21,0.49,0.74}
\usepackage[pagebackref,breaklinks,colorlinks,allcolors=cvprblue]{hyperref}
\usepackage{booktabs}
\usepackage{multirow}
\usepackage{graphicx, amsmath, bbm, soul}

\def\paperID{9651} 
\def\confName{CVPR}
\def\confYear{2025}

\title{SemiDAViL: \underline{Semi}-supervised \underline{D}omain \underline{A}daptation with \underline{Vi}sion-\underline{L}anguage Guidance for Semantic Segmentation}

\author{Hritam Basak*, Zhaozheng Yin\\
Dept. of Computer Science, Stony Brook University, NY, USA\\
{\tt\small \textsuperscript{*}hbasak@cs.stonybrook.edu}
}

\begin{document}
\maketitle
\input{sec/0_abstract}    
\input{sec/1_intro}
\input{sec/2_related_works}
\input{sec/3_method}
\input{sec/4_results}

\input{sec/5_conclusion}

\input{sec/6_acknowledgement}

\renewcommand{\thesection}{\Alph{section}}
\setcounter{section}{0}

\begin{center}
    \twocolumn[\centering\Large\bf Supplementary Material\vspace{0.5cm}]
\end{center}

\input{sec/X_suppl}
{
    \small
    \bibliographystyle{ieeenat_fullname}
    \bibliography{main}
}

\end{document}

%% file: sec/0_abstract.tex
\begin{abstract}
Domain Adaptation (DA) and Semi-supervised Learning (SSL) converge in Semi-supervised Domain Adaptation (SSDA), where the objective is to transfer knowledge from a source domain to a target domain using a combination of limited labeled target samples and abundant unlabeled target data. Although intuitive, a simple amalgamation of DA and SSL is suboptimal in semantic segmentation due to two major reasons: (1) previous methods, while able to learn good segmentation boundaries, are prone to confuse classes with similar visual appearance due to limited supervision; and (2) skewed and imbalanced training data distribution preferring source representation learning whereas impeding from exploring limited information about tailed classes. Language guidance can serve as a pivotal semantic bridge, facilitating robust class discrimination and mitigating visual ambiguities by leveraging the rich semantic relationships encoded in pre-trained language models to enhance feature representations across domains. Therefore, we propose the first language-guided SSDA setting for semantic segmentation in this work. Specifically, we harness the semantic generalization capabilities inherent in vision-language models (VLMs) to establish a synergistic framework within the SSDA paradigm. To address the inherent class-imbalance challenges in long-tailed distributions, we introduce class-balanced segmentation loss formulations that effectively regularize the learning process. Through extensive experimentation across diverse domain adaptation scenarios, our approach demonstrates substantial performance improvements over contemporary state-of-the-art (SoTA) methodologies. Code is available: \href{https://github.com/hritam-98/SemiDAViL}{GitHub}.
\end{abstract}

%% file: sec/1_intro.tex
\vspace{-5mm}
\section{Introduction}
\label{sec:intro}
\vspace{-2mm}
\begin{figure}
    \centering
    \includegraphics[width=\linewidth]{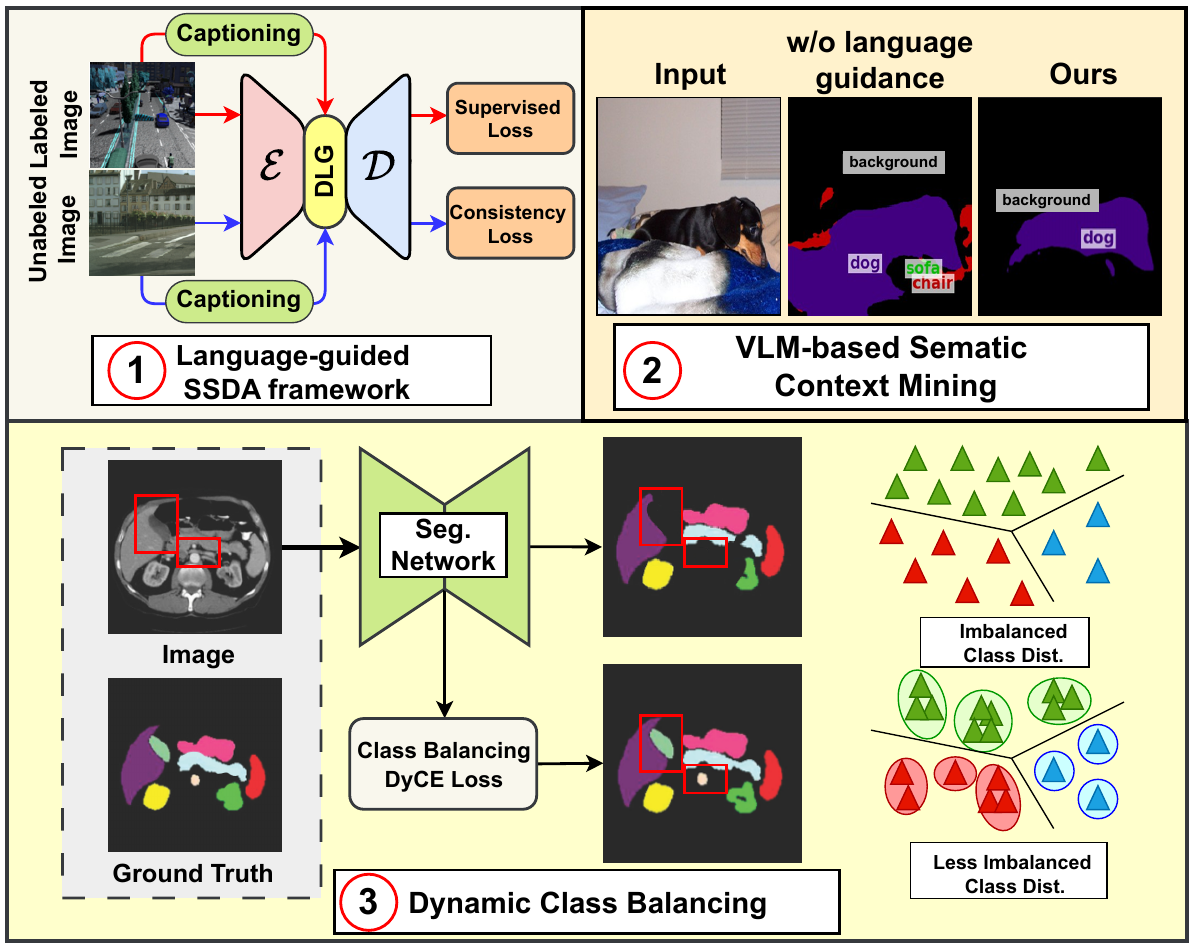}
    \caption{Major contributions of SemiDAVil: (1) We propose the first language-guided SSDA framework for semantic segmentation, (2) Utilizing spatial context via dense language guidance (DLG) improves segmentation performance, (3) Our proposed DyCE loss dynamically reweighs imbalanced class distributions, resulting in precise segmentation of \textcolor{red}{minority classes}. }
    \vspace{-5mm}
    \label{fig:teaser}
\end{figure}
The remarkable progress in deep learning has significantly enhanced the performance of visual understanding tasks, including image classification \cite{chen2021review}, object detection \cite{zou2023object}, and, more recently, semantic segmentation \cite{ren2023visual,thisanke2023semantic}. These advancements have been particularly notable when a wealth of labeled training data is available. However, as noted in \cite{shen2023survey}, their performance degrades precipitously when confronted with annotation-scarce environments, especially in the context of semantic segmentation, where dense pixel-wise annotations are essential. Furthermore, these sophisticated models exhibit substantial vulnerability when tasked with generalizing across domains characterized by significant distributional shifts \cite{kang2020pixel,hoyer2022hrda} - a challenge particularly evident in real-world applications where models trained on synthetic data must maintain robust performance in naturalistic settings, such as autonomous navigation systems \cite{schwonberg2023survey,erkent2020semantic}. This inherent limitation in cross-domain generalization has catalyzed the emergence of two pivotal research paradigms: Domain Adaptation (DA) and Semi-supervised Learning (SSL).

The confluence of DA and SSL has given rise to Semi-supervised Domain Adaptation (SSDA), a hybrid approach that strategically leverages three distinct data streams: comprehensively labeled source domain data, sparsely labeled target domain samples, and a wealth of unlabeled target domain instances \cite{saito2019semi, li2021cross}. While SSDA holds intuitive appeal for real-world applications, existing methods encounter critical limitations when applied to semantic segmentation tasks. Specifically, \textbf{(1)} despite achieving accurate segmentation boundaries, current approaches \cite{yang2023revisiting, ma2023enhanced} often suffer from \underline{\textit{misclassification among visually similar classes}}, due to restricted supervision within the target domain; \textbf{(2)} the SSDA framework tends to over-prioritize source domain features, driven by abundant source labels, while generating \underline{\textit{error-prone pseudo-labels for target data}}, which hampers adaptation performance; \textbf{(3)} class-imbalance, a common issue in real-world datasets, exacerbates these challenges, limiting effective exploration and representation of \underline{\textit{minority (tail) classes}} in the target domain.

To address the identified SSDA challenges, we augment the SSDA paradigm with vision-language (VL) guidance using VLMs (e.g., CLIP \cite{radford2021learning}) to enrich semantic representation, leveraging their large-scale image-caption pretraining. By incorporating VLM features into a global-local context exploration module, we \underline{mitigate misclassification among visually similar classes}. To tackle the over-reliance on source features, we introduce a joint embedding space guided by language priors, enhancing instance separability and \underline{reducing domain bias}, unlike traditional divergence-based alignment methods \cite{yu2023high,li2020maximum}. Finally, to \underline{combat class imbalance}, we design a tailored cross-entropy loss that dynamically reweighs minority classes, thereby facilitating more equitable exploration and representation of tail classes in the target domain. Specifically, our contributions can be summarized as:
\begin{enumerate}
    \item \textbf{Language-Guided SSDA Framework}: We pioneer the first language-guided SSDA framework for semantic segmentation by harnessing the rich semantic knowledge encoded in pre-trained Vision-Language Models (VLMs). Our novel attention-based fusion mechanism seamlessly integrates visual features with dense language embeddings, establishing a robust semantic bridge between source-target domains while providing enhanced contextual understanding.
    \item \textbf{Enhanced Feature Localization}: Recognizing that VL pre-training primarily operates at the image level, we address the critical challenge of feature localization in semantic segmentation through targeted fine-tuning. To mitigate the risks of overfitting and semantic knowledge degradation inherent in limited-annotation scenarios, we develop a sophisticated consistency regularization framework that preserves the rich semantic representations acquired during pre-training.
    \item \textbf{Adaptive Class-Balanced Loss}: To tackle class imbalance in a limited annotation scenario, we introduce a Dynamic Cross-Entropy (DyCE) loss formulation that dynamically calibrates the learning emphasis on tail classes. This innovative, plug-and-play loss mechanism demonstrates broad applicability across various class-imbalanced learning scenarios.
    \item \textbf{State-of-the-Art Performance}: Through detailed evaluation across diverse domain-adaptive and class-imbalanced segmentation benchmarks, our methodology demonstrates superior performance and robustness, consistently surpassing contemporary state-of-the-art approaches by significant margins.
\end{enumerate}

%% file: sec/2_related_works.tex
\section{Related Works}\label{sec:related_works}

\subsection{Semi-supervised Domain Adaptation}\label{subsection:semi-supervised-learning}
Recent advances in Semi-Supervised Domain Adaptation (SSDA) for semantic segmentation have focused on utilizing limited labeled target data and abundant unlabeled data to bridge the domain gap at the pixel level \cite{basak2024quest,basak2025forget}. Early approaches like MME \cite{saito2019semi} and ASDA \cite{qin2022semi} used entropy minimization for feature alignment, but their classification-centric strategies struggled with fine-grained segmentation tasks, leading to suboptimal boundary delineation. To address this, SSL-based methods such as DECOTA \cite{yang2021deep} and SS-ADA \cite{yan2024ss} employed teacher-student frameworks with consistency constraints, generating pseudo-labels for unlabeled target data. However, these methods faced issues with noisy pseudo-labels, particularly for minority and boundary classes. 
More recent methods have explored novel directions: S-Depth \cite{hoyer2023improving} leverages self-supervised depth estimation as an auxiliary task to enhance feature learning, while DSTC \cite{gao2024delve} introduces a domain-specific teacher-student framework that dynamically adapts to target domain characteristics. IIDM \cite{fu2024iidminterintradomainmixing} proposes an innovative inter-intra-domain mixing strategy to address domain shift and limited supervision simultaneously. However, these methods often struggle with two critical limitations: class confusion due to limited supervision and skewed data distribution favoring source domain representations. 

\subsection{Vision Language Model}\label{subsection:vlm-related-works}

Vision-Language Models (VLMs), like CLIP and its extensions \cite{radford2021learning, ilharco, lueddecke22_cvpr, zhou2023non}, leverage large-scale image-text pre-training for semantic segmentation via a shared embedding space that aligns visual and textual features. Initial zero-shot methods, such as MaskCLIP \cite{zhou2022extract} and GroupViT \cite{xu2022groupvit}, struggled with boundary precision due to reliance on high-level features. Later, fine-tuned models like OpenSeg \cite{ghiasi2022scaling} and LSeg \cite{lilanguage} improved segmentation accuracy using labeled data and text embeddings. Techniques such as ZegFormer \cite{ding2022decoupling} and OVSeg \cite{liang2023open} utilize frozen CLIP features for mask proposal classification, while ZegCLIP \cite{zhou2023zegclip} aligns dense visual-text embeddings in a streamlined manner.
Recently, SemiVL \cite{hoyer2025semivl} has shown that language cues can enhance semantic insights and mitigate class confusion; however, their application in domain adaptation remains under-explored. The recent LIDAPS model \cite{mansour2024language} and follow-up works \cite{fahes2024simple, kim2024lc, wu2024clip2uda} apply language guidance for domain bridging in panoptic segmentation but rely on manual thresholding for pseudo-mask filtering and a complex multi-stage training process. 
Despite their improvements, these methods still misclassify tail classes (e.g., \textit{fence, bike, wall}), posing critical risks for applications like autonomous driving, where errors in identifying such objects can lead to severe consequences.

\subsection{Class-Imbalance Handling}\label{class-imbalance-literature}

Class imbalance significantly hinders real-world semantic segmentation, as small object classes often appear less frequently and cover fewer pixels than dominant background classes, unlike balanced datasets like CIFAR-10/100, ImageNet, and Caltech-101/256 \cite{krizhevsky2010cifar, deng2009imagenet, griffin_holub_perona_2022}. Data-level methods like oversampling/undersampling adjust sampling probabilities for minority classes but struggle in dense tasks due to uneven class distribution \cite{zhou2023dynamic, saini2023tackling}. Algorithmic strategies such as class-weighted losses address bias by penalizing rare classes more \cite{lin2017focal}, but treating small object classes equally often leads to instability \cite{yan2019weighted, sudre2017generalised}.

In unsupervised domain adaptation (UDA) for segmentation, common strategies include data-level adjustments using source domain frequencies \cite{hoyer2022daformer, hoyer2022hrda, hoyer2023mic}, and adaptive weighting based on target statistics \cite{yan2019weighted}, but these are computationally costly for dense predictions. Approaches that relax pseudo-label filtering for rare classes still inherit source biases, causing misclassifications \cite{zou2018unsupervised}. Most UDA methods prioritize data-level sampling \cite{muhammad2022vision}, overlooking the synergy of combining data and algorithmic approaches, which remains impractical and lacks generalizability for diverse tasks \cite{shen2023survey, wu98gradient, wang2024towards}.

%% file: sec/3_method.tex
\section{Proposed Method}\label{sec:method}

\begin{figure}[t!]
    \centering
    \includegraphics[width=\columnwidth]{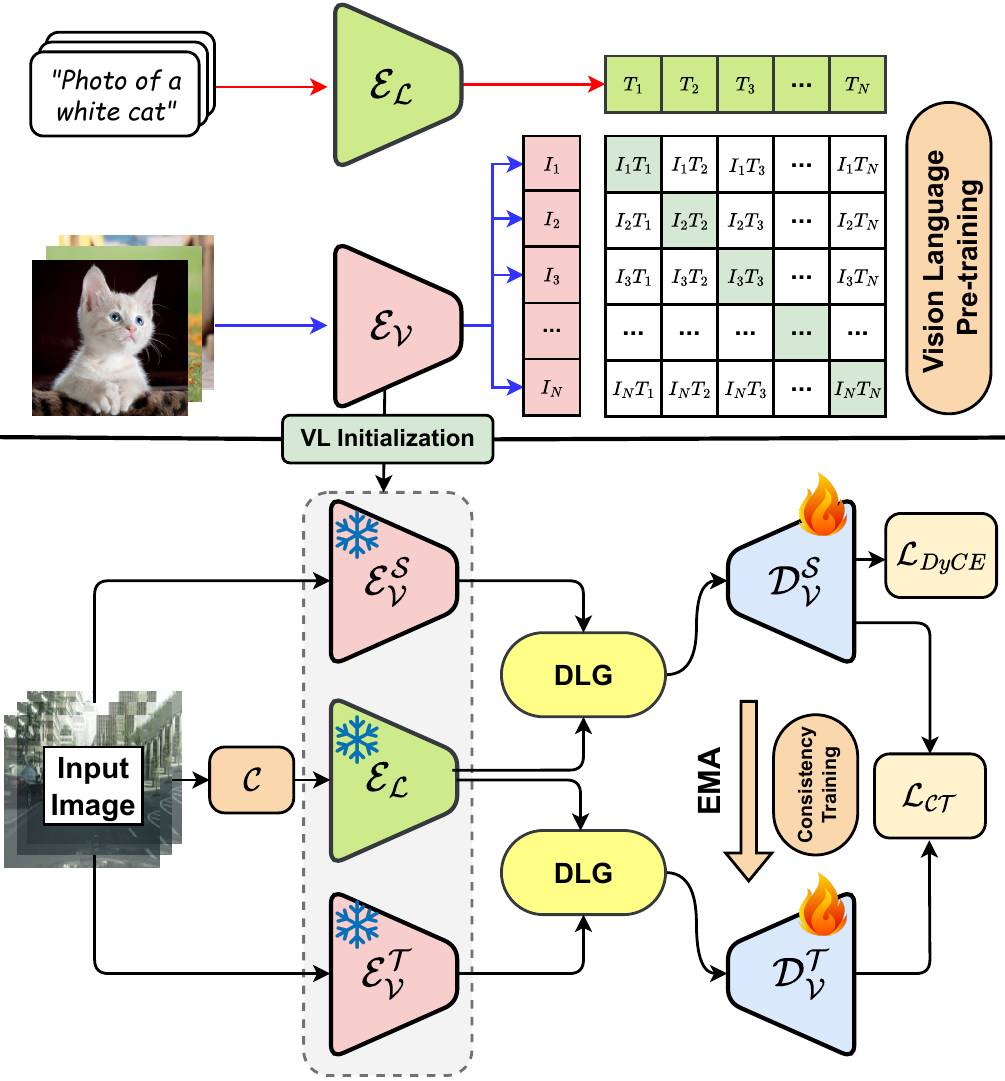}
    \caption{Overview of \textbf{SemiDAViL}: We leverage Vision-Language (VL) Pre-training (top) to initialize the language encoder \(\mathcal{E}_\mathcal{L}\) and vision encoders $\mathcal{E}_\mathcal{V}^{\{\mathcal{S, T}\}}$ in a semi-supervised setting (bottom), where \(\mathcal{S}\) and \(\mathcal{T}\) denote the student and teacher branches, respectively. To bridge image-level VL features for dense pixel-level tasks, we utilize a captioning model \(\mathcal{C}\) to generate text descriptions of images and a Dense Language Guidance (DLG) module. The framework is trained with a supervised loss \(\mathcal{L}_{DyCE}\) for labeled data and a consistency loss \(\mathcal{L}_{\mathcal{CT}}\) for unlabeled data.}
    \label{fig:overall}
    \vspace{-5mm}
\end{figure}

In our SSDA setting, we utilize image-label pair from the source domain \(\mathcal{D}^{\mathcal{S}r}\): \(\{(x^{\mathcal{S}r}_i, y^{\mathcal{S}r}_i)\}_{i=1}^{\mathbb{N}^{\mathcal{S}r}}\), a limited set of labeled target samples \(\{(x^{{T}r_\mathcal{L}}_i, y^{Tr_\mathcal{L}}_i)\}_{i=1}^{\mathbb{N}^{Tr_\mathcal{L}}}\), and a large pool of unlabeled target data \(\{(x^{Tr_\mathcal{U}}_i)\}_{i=1}^{\mathbb{N}^{Tr_\mathcal{U}}}\), where \(\mathbb{N}^{Tr_\mathcal{U}} \gg \mathbb{N}^{Tr_\mathcal{L}}\). 
Our proposed SemiDAViL framework effectively tackles the challenges of SSDA by leveraging VL-pretrained encoders (\autoref{subsection-vlm-pretrain}) for enriched semantic representation learning from ${\mathcal{S}r}\cup Tr_\mathcal{L}\cup Tr_\mathcal{U}$, addressing the issue of misclassification among visually similar classes. We incorporate dense semantic guidance from language embeddings (\autoref{subsection:dense-language-guidance}) to enhance instance separability and reduce domain bias. Consistency-regularized SSL (\autoref{subsection:consistency-training}) mitigates over-reliance on source features, while the class-balancing DyCE loss (\autoref{subsection:class-balanced-loss}) combats class imbalance by reweighting tail classes. The overall architecture of SemiDAViL is outlined in \autoref{fig:overall}.

\subsection{Vision-Language Pre-training}\label{subsection-vlm-pretrain}
Previous regularization-based SSDA methods have shown effectiveness in semi-supervised semantic segmentation by enforcing stable predictions on unlabeled data. However, as discussed in \autoref{sec:related_works}, they often struggle with distinguishing visually similar classes, especially when only a limited set of labeled target samples \(\{(x^{\mathcal{T_L}}_i, y^{\mathcal{T_L}}_i)\}\) is available. The primary issue arises due to the lack of diverse semantic coverage, leading to errors in class discrimination. To address this, we leverage Vision-Language Models (VLMs) like CLIP \cite{radford2021learning}, which are trained on large-scale image-text datasets, \(\mathcal{D}_{\text{clip}} = \{(x, t)\}\), where \(x\) and \(t\) are images and their associated captions. CLIP consists of a vision encoder \(\mathcal{E}_\mathcal{V}\) and a language encoder \(\mathcal{E}_\mathcal{L}\), optimized jointly using a contrastive loss:
\begin{equation}
\mathcal{L}_{\text{contrast}} = -\sum_{i=1}^{N} \log \frac{\exp(\langle \mathcal{E}_\mathcal{V}(x_i), \mathcal{E}_\mathcal{L}(t_i) \rangle / \tau)}{\sum_{j=1}^{N} \exp(\langle \mathcal{E}_\mathcal{V}(x_i), \mathcal{E}_\mathcal{L}(t_j) \rangle / \tau)},    
\end{equation}
where \(\langle \cdot, \cdot \rangle\) denotes the cosine similarity and \(\tau\) is a temperature parameter. This objective aligns the visual and textual embeddings into a shared semantic space, learning robust, class-agnostic representations that generalize across diverse classes. Unlike conventional ImageNet pre-training that relies on manually annotated labels, CLIP’s training with web-crawled image-caption pairs allows it to capture richer semantics without restricting to a fixed set of categories.

To mitigate the limited semantic knowledge in standard consistency training in our SSDA framework, we initialize our (student-teacher) segmentation encoders \(\mathcal{E}_\mathcal{V}^{\{\mathcal{S, T}\}}\) with CLIP’s pre-trained vision encoder \(\mathcal{E}_\mathcal{V}\), rather than using an ImageNet-trained backbone. This transfer of rich semantic priors enables enhanced feature extraction and better semantic differentiation (as found in \autoref{tab:ablation} and well supported in \cite{hoyer2025semivl}), particularly for visually ambiguous classes, leading to more robust segmentation performance.

\begin{figure}[t]
    \centering
    \includegraphics[width=0.75\linewidth]{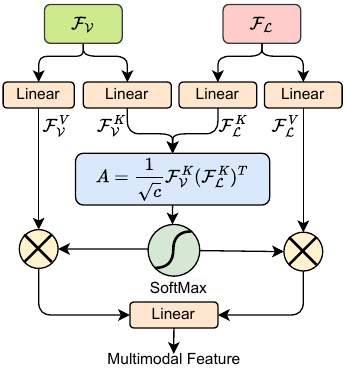}
    \caption{Overall architecture of our proposed DLG module: it is based on dense similarity maps of the vision and text embeddings. More details are provided in \autoref{subsection:dense-language-guidance}.}
    \label{fig:attention}
    \vspace{-5mm}
\end{figure}

\subsection{Dense Language Guidance (DLG)}\label{subsection:dense-language-guidance}
Most prior VLM methods employ a standard attention mechanism for multi-modal feature integration \cite{ding2021vision, zhang2023cross,liu2023multi}, i.e., features from two modalities (query and key) generate an attention matrix to aggregate vision features based on language-derived weights. However, this approach only utilizes the language feature to compute attention scores, without directly incorporating it into the fused output, effectively treating the result as a reorganized single-modal vision feature. Consequently, the output vision feature dominates the decoder, leading to a substantial loss of language information. 
Based on our empirical findings (provided in \textbf{supplementary file}, and well supported in \cite{liu2023multi}), we argue that while generic attention effectively processes value inputs, it fails to fully exploit query features for deep cross-modal interaction, resulting in insufficient fusion of vision and language modalities.

To address this, we utilize Dense Language Guidance (DLG) that transforms both the vision and language features into key-query pairs and treats them equally, as shown in \autoref{fig:attention}. First, visual features $\mathcal{F}_\mathcal{V}\in\mathbb{R}^{h\times w\times c}$ with $h\times w$ dimension and $c$ channels for image $\mathcal{X}$ are extracted through $\mathcal{E}_\mathcal{V}$\footnote{student-teacher encoders $\mathcal{E}_\mathcal{V}^{\{\mathcal{S,T}\}}$ represented as $\mathcal{E}_\mathcal{V}$ for simplicity.} with frozen weight $\mathcal{\phi_\mathcal{V}}$: $\mathcal{F}_\mathcal{V}\leftarrow\mathcal{E}_\mathcal{V}(\mathcal{X};\phi_\mathcal{V}^{\includegraphics[width=01em]{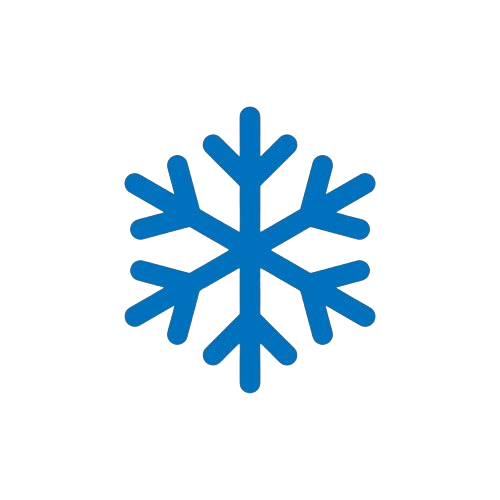}})$. To further utilize language embeddings $\mathcal{F}_\mathcal{L}\in \mathbb{R}^{n_\mathcal{L}\times c}$, we extract text description with $n_{\mathcal{L}}$ tokens for $\mathcal{X}$ using off-the-shelf captioning model $\mathcal{C}$, followed by CLIP-initialized language encoder $\mathcal{\mathcal{E}_\mathcal{L}}$ with frozen weights $\phi_\mathcal{L}$: $\mathcal{F}_\mathcal{L}\leftarrow\mathcal{E}_\mathcal{L}\big(\mathcal{C}(\mathcal{X});\phi_\mathcal{L}^{\includegraphics[width=1em]{freeze.png}}\big)$. This is followed by projection of $\mathcal{F}_{\{\mathcal{V,L}\}}$ to key-value pairs using linear layers: $\mathcal{F}_{\{\mathcal{V,L}\}}^{\{K,V\}}\leftarrow \texttt{Linear}(\mathcal{F}_{\{\mathcal{V,L}\}})$. Next, multi-modal key values are used to generate an attention matrix $\mathcal{A}\in\mathbb{R}^{n_\mathcal{L}\times h\times w}$:
\begin{equation}
    \mathcal{A}=\frac{1}{\sqrt{c}}\mathcal{F}_\mathcal{V}^K\cdot\big(\mathcal{F}_\mathcal{L}^K\big)^T
\end{equation} 
Instead of applying attention to a single modality as in conventional methods, we normalize across both dimensions and compute cross-attention on vision and language features. Specifically, we employ a SoftMax activation followed by attention over \(\mathcal{F}_\mathcal{V}^V\) and \(\mathcal{F}_\mathcal{L}^V\) to generate language-attended vision features and vision-attended language features, respectively, ensuring balanced and comprehensive feature fusion:
\begin{equation}
\begin{aligned}
     \mathcal{F}_\mathcal{V}^\mathcal{A}=\texttt{SoftMax} [\mathcal{A}]\mathcal{F}_\mathcal{V}^V \\
    \mathcal{F}_\mathcal{L}^\mathcal{A}=\texttt{SoftMax} [\mathcal{A}]\mathcal{F}_\mathcal{L}^V    
\end{aligned}
\end{equation}
Finally, these two attended feature maps are combined to generate a true multimodal feature representation $\mathcal{F}_\mathcal{M}\in\mathbb{R}^{n_\mathcal{L}\times h\times w}$: $\mathcal{F}_\mathcal{M}=\mathcal{F}_\mathcal{V}^\mathcal{A}\cdot\big(\mathcal{F}_\mathcal{L}^\mathcal{A}\big)^T$, which is the basis of our VL-guided SSDA pipeline. This is thereafter passed to the student-teacher decoders $\mathcal{D}_\mathcal{V}^{\{\mathcal{S,T}\}}$ for consistency training.

\subsection{Consistency Training (CT)}\label{subsection:consistency-training}
To effectively utilize labeled and unlabeled data in our SSDA setting, we utilize a student-teacher network \cite{tarvainen2017mean} for consistency training. Specifically, for unlabeled target data $\{(x^{Tr_\mathcal{U}}_i)\}_{i=1}^{\mathbb{N}^{Tr_\mathcal{U}}}\in\mathcal{D}^{Tr_\mathcal{U}}$, we obtain two multimodal features $\mathcal{F_M^S, F_M^T}$ (from DLG), and pass them through two identical but differently initialized decoders $\{\mathcal{D_V^S, D_V^T}\}$ with trainable parameters $\{\theta_\mathcal{V}^{\mathcal{S}\includegraphics[width=1em]{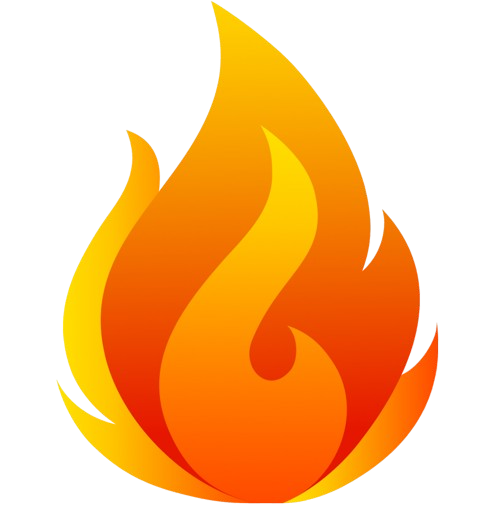}}, \theta_\mathcal{V}^{\mathcal{T}\includegraphics[width=1em]{fire.png}}\}$, respectively, and enforce their predictions to be consistent:
\begin{equation}
\begin{aligned}
        L_{CT}= \frac{1}{\mathbb{N}^{Tr_\mathcal{U}}} \sum\limits_{x_i\in \mathcal{D}^{\mathcal{T_U}}}
    \sum\limits_{p=1}^{h\times w} \underset{max({y}^\mathcal{S}_p)\geq Th}{\mathbbm{1}} \: CE \big( y^\mathcal{S}_p, y^\mathcal{T}_p   \big)
\end{aligned}
\end{equation}
where $y^\mathcal{S}_p$ and $y^\mathcal{T}_p$ are the $p^{\text{th}}$ pixel prediction from student and teacher model: $y^{m}\leftarrow \mathcal{D}_\mathcal{V}^{m}(\mathcal{F}_\mathcal{M}^{m};\theta_\mathcal{V}^{m}); m=\{\mathcal{S,T}\}$, $Th$ is a threshold to exclude noisy pseudo-labels in $L_{CT}$. To further utilize labeled target data $\{\mathcal{D}^{\mathcal{S}r}\cup\mathcal{D}^{Tr_\mathcal{L}}\}$, we can employ a supervised CE loss between ground truth $y$ and student prediction $y^\mathcal{S}$:
\begin{equation}
    L_{S}=
    \frac{1}{\mathbb{N}^{\mathcal{S}r}+\mathbb{N}^{Tr_\mathcal{L}}} \sum\limits_{x_i\in \{\mathcal{D}^{\mathcal{S}r}\cup\mathcal{D}^{Tr_\mathcal{L}}\}}
    \sum\limits_{c=0}^{N_C} CE\big(y_c^\mathcal{S}, y_c\big),
\end{equation}
where $N_C$ represents the number of classes. However, $L_{S}$ might incur suboptimal performance due to inherent class imbalance, as discussed in \autoref{class-imbalance-literature} and evident in previous methods in \autoref{tab:class-wise-comparison}, \autoref{tab:medical-comparison}. We propose a dynamic CE loss to alleviate this shortcoming, as detailed in \autoref{subsection:class-balanced-loss}. The student branch is updated using a combined consistency and DyCE loss, whereas the teacher model is updated using an exponential moving average (EMA) of the student parameters:
\begin{equation}
    \theta^\mathcal{T}_\mathcal{V} (t) \leftarrow \alpha\theta^\mathcal{T}_\mathcal{V}(t-1) + (1-\alpha)\theta^\mathcal{S}_\mathcal{V}(t)
\end{equation}
where $t$ is step number, $\alpha$ is the momentum coefficient \cite{he2020momentum}. 

\subsection{Class-balanced Dynamic CE Loss (DyCE)}\label{subsection:class-balanced-loss}
The Cross-Entropy (CE) loss measures the difference between predicted probabilities and ground truth labels by computing a negative log-likelihood for each class, averaged over all instances in the mini-batch:
\begin{equation}
   {L}_{\text{CE}} = -\frac{1}{S} \sum_{i=1}^{S} \sum_{c=0}^{N_C} y_{i,c} \log p_{i,c}, 
\end{equation}
where \(y_{i,c}\) and \(p_{i,c}\) are the GT and predicted probability for class $c$, \(S\) is the batch size. Taking the gradient of $L_{CE}$ for each instance, we have:
\begin{equation}\label{original-CE-gradient}
    \frac{\partial \mathcal{L}_{\text{CE}}}{\partial p_{i,c}} = 
\begin{cases}
-\frac{1}{S} \frac{1}{p_{i,c}}, & \text{if } y_{i,c} = 1, \\
0, & \text{otherwise}.
\end{cases}
\end{equation}
Hence, CE loss only updates the gradient for the target class per instance, using a uniform weight of \(-\frac{1}{S}\). This leads to two key problems in large imbalanced datasets: (1) equal weighting across classes overlooks class imbalance, treating frequent and rare classes the same; (2) the gradient magnitude becomes vanishingly small as \(N\) scales to millions, causing ineffective updates (gradient vanishing). While recent studies have proposed reweighting schemes to address the class imbalance issue \cite{qiu2023subclassified,truong2023fredom}, they fail to tackle the core problem of diminished gradient magnitudes (refer to supporting evidence in \textbf{supplementary file}), limiting the optimization efficiency in dense segmentation tasks.

To address this, we propose a Dynamic CE (DyCE) loss that dynamically adjusts the weighting of gradients based on the class distribution within each mini-batch, addressing the persistent class imbalance issue that remains even after discarding simple instances. The key idea is to adaptively align the gradient contributions to the real-time class distribution at every training step. This is formalized as:
\begin{equation}
    {L}_{\text{DyCE}} = -\frac{1}{f_H^{\omega}} \sum_{c=0}^{N_C} \frac{1}{f_c^{(1-\omega)}} \sum_{i=1}^S \underset{i\in H}{\mathbbm{1}} y_{i,c} \:\: \log p_{i,c},    
\end{equation}
where \(f_c = \sum_{i \in H} y_{i,c}\) is the total count of class \(c\) in the mined subset \(H\) which consists of $h\%$ hardest instances from the batch, $f_H=|H|$ is the count of instances in subset $H$. The loss computation involves four key steps: (1) computing the standard CE loss for each sample; (2) creating a subset $H$ from the batch, similar to \cite{wu2016bridging}; (3) assigning dynamic class weights \(\frac{1}{f_c^{(1-\omega)}}\), inversely proportional to the mined class frequency; and (4) scaling the loss by a volume weight \(-\frac{1}{f_H^{\omega}}\), which adjusts for the batch size and mined subset size. The hyperparameter \(\omega \in (0, 1)\) acts as a weight-balancing factor, balancing the influence of instance-level and class-level weighting. The gradient of DyCE loss is:
\begin{equation}
    \frac{\partial L_{DyCE}}{\partial p_{i,c}} = \begin{cases} 
-\frac{1}{f_H^\omega}  \frac{1}{f_c^{(1-\omega)}}  p_{i,c} & \text{if } y_{i,c} = 1, \\
0 & \text{otherwise.}
\end{cases}    
\end{equation}
Here $\frac{1}{f_H^\omega}  \frac{1}{f_c^{(1-\omega)}}\geq \frac{1}{S}$ as compared to \autoref{original-CE-gradient}, as $S\geq f_H\geq f_c$ and hence the vanishing gradient issue is resolved. 


%% file: sec/4_results.tex
\section{Experimental Results}\label{sec:results}

\subsection{Dataset Description}\label{dataset}
We evaluate our proposed SSDA method on a segmentation task by adapting from two synthetic datasets, GTA5 \cite{richter2016playing} and SYNTHIA \cite{ros2016synthia}, to the real-world Cityscapes dataset \cite{cordts2016cityscapes}. The Cityscapes dataset consists of 2,975 training images and 500 validation images, all manually annotated with 19 classes. Since the test set annotations are not publicly available, we evaluate on the validation set, and tune the hyper-parameters on a small subset of the training set, following previous works \cite{cardace2022shallow,hoyer2023improving}. GTA5 provides 24,966 training images, and we consider the 19 classes that overlap with Cityscapes. The SYNTHIA dataset includes 9,400 fully labeled images, and we evaluate results based on the 16 classes it shares with Cityscapes. 

Furthermore, to validate our DyCE loss's effectiveness, we evaluate on an extremely imbalanced medical dataset, Synapse  \cite{landman2015miccai}. The Synapse dataset comprises 30 CT scans covering 13 different organs (i.e., foreground classes): spleen (Sp), right and left kidneys (RK/LK), gallbladder (Ga), esophagus (Es), liver (Li), stomach (St), aorta (Ao), inferior vena cava (IVC), portal and splenic veins (PSV), pancreas (Pa), and right and left adrenal glands (RAG/LAG). In this dataset, foreground voxels make up only 4.37\% of the entire dataset, with 95.63\% background, and the right adrenal gland contributes a mere 0.14\% of foreground, whereas liver consists of 53.98\% foreground, underscoring the severe class imbalance. Following the setup of \cite{wang2024towards}, we split the dataset into 20 scans for training, 4 for validation, and 6 for testing. 
\vspace{-2mm}

\subsection{Implementation Details}\label{subsection:implementation}
Following SemiVL \cite{hoyer2025semivl}, we utilize ViT-B/16 vision encoder \cite{dosovitskiy2020vit} and a Transformer text encoder \cite{vaswani2017attention}, both initialized with CLIP pre-training \cite{radford2021learning} and generate dense embeddings following \cite{zhou2022extract}. The initial learning rate is set to \(10^{-4}\), decaying exponentially with a factor of \(0.9\). We set the weight decay to \(2 \times 10^{-4}\) and momentum to \(0.9\). Following \cite{kalluri2024tell}, we use BLIP-2 \cite{li2023blip} as our off-the-shelf captioning model $\mathcal{C}$ for all domains. $Th$, $\alpha$ is set to $0.95, 0.999$, following \cite{hoyer2025semivl, he2020momentum}. Following \cite{fu2024iidminterintradomainmixing,tranheden2021dacs}, source images are resized to \(760 \times 1280\) and target images to \(512 \times 1024\), followed by random cropping to \(512 \times 512\). SemiDAViL is trained for 40k iterations which takes $\sim15$ hours on an NVIDIA RTX4090 GPU using Python environment. 
\subsection{Findings and Comparison with SoTA}\label{comparison-with-sota}
\begin{table}[tbp]
\centering
\caption{Quantitative comparison of our proposed method with existing unsupervised domain adaptation (UDA), semi-supervised learning (SSL), and semi-supervised domain adaptation (SSDA) methods on \textbf{GTA5 $\to$ Cityscapes} benchmark. We report 19-class mIoU scores on the Cityscapes validation set across 0, 100, 200, 500, 100, and 2975 (100\%) labeled target images. Our results are \textbf{highlighted} whereas the previous best and second-best results are marked in \textcolor{red}{red} and \textcolor{blue}{blue}.  }
\vspace{-2mm}
\label{tab:comparison-GTA}
\resizebox{\columnwidth}{!}{%
\begin{tabular}{@{}cccccccc@{}}
\toprule
 &  & \multicolumn{6}{c}{\textbf{Labeled Target Samples}} \\ \cmidrule(l){3-8} 
\multirow{-2}{*}{\textbf{Type}} & \multirow{-2}{*}{\textbf{Methods}} & \textbf{0} & \textbf{100} & \textbf{200} & \textbf{500} & \textbf{1000} & \textbf{2975} \\ \midrule
Supervised & DeepLabV2 \cite{chen2017deeplab} & - & 43.0 & 48.3 & 54.8 & 58.3 & 66.1 \\ \midrule
 & PRoDA \cite{zhang2021prototypical} & 57.5 & - & - & - & - & - \\
 & DaFormer \cite{hoyer2022daformer}& 56.0 & - & - & - & - & - \\
 & CONFETI \cite{li2023contrast} & {\color[HTML]{3531FF} 62.2} & - & - & - & - & - \\
\multirow{-4}{*}{UDA} & DIGA \cite{shen2023diga}& {\color[HTML]{FE0000} 62.7} & - & - & - & - & - \\ \midrule
 & CowMix \cite{french2019semi}& - & 50.8 & 54.8 & 61.7 & 64.8 & - \\
 & ClassMix \cite{olsson2021classmix}& - & 54.4 & 58.6 & 62.1 & 64.3 & - \\
 & CPS (S) \cite{chen2021semi}& - & 55.0 & 59.5 & 63.0 & 65.7 & - \\
 & CPS (E) \cite{chen2021semi}& - & {\color[HTML]{3531FF} 55.3} & {\color[HTML]{3531FF} 60.0} & {\color[HTML]{3531FF} 63.6} & {\color[HTML]{3531FF} 66.3} & - \\
\multirow{-5}{*}{SSL} & DusPerb \cite{yang2023revisiting}& - & {\color[HTML]{FE0000} 61.8} & {\color[HTML]{FE0000} 66.7} & {\color[HTML]{FE0000} 68.4} & {\color[HTML]{FE0000} 72.1} & - \\ \midrule
 & ASS \cite{xie2022towards}& - & 54.2 & 56.0 & 60.2 & 64.5 & 69.1 \\
 & DECOTA (S) \cite{yang2021deep}& - & 60.7 & 61.8 & 64.2 & 66.3 & 69.2 \\
 & DECOTA (E) \cite{yang2021deep}& - & 61.3 & 62.3 & 64.7 & 67.0 & 69.9 \\
 & DLDM \cite{chen2021semidual}& - & 61.2 & 60.5 & 64.3 & 66.6 & 69.8 \\
 & SSDDA \cite{chen2022semi}& - & 60.1 & 62.9 & 65.7 & 66.8 & - \\
 & DSTC (S) \cite{gao2024delve}& - & 64.5 & 65.8 & 69.2 & 70.3 & 71.9 \\
 & DSTC (E) \cite{gao2024delve}& - & 65.2 & 66.4 & {\color[HTML]{3531FF} 70.0} & {\color[HTML]{3531FF} 70.9} & {\color[HTML]{3531FF} 72.6} \\
 & S-Depth \cite{hoyer2023improving}& - & {\color[HTML]{3531FF} 66.1} & {\color[HTML]{3531FF} 67.3} & 69.9 & 70.5 & 71.7 \\
 & IIDM \cite{fu2024iidminterintradomainmixing}& - & {\color[HTML]{FE0000} 69.5} & {\color[HTML]{FE0000} 70.0} & {\color[HTML]{FE0000} 70.6} & {\color[HTML]{FE0000} 72.8} & {\color[HTML]{FE0000} 73.3} \\ \cmidrule(l){2-8} 
 & \textbf{Ours w/o DyCE} & \textbf{66.9} & \textbf{70.3} & \textbf{71.8} & \textbf{72.1} & \textbf{73.9} & \textbf{74.7} \\ \cmidrule(l){2-8} 
\multirow{-11}{*}{SSDA} & \textbf{Ours w/ DyCE} & \textbf{67.7} & \textbf{71.1} & \textbf{72.5} & \textbf{72.9} & \textbf{74.8} & \textbf{75.2} \\ \bottomrule
\end{tabular}%
}
\vspace{-7mm}
\end{table}

\begin{table}[tbp]
\centering
\caption{Quantitative comparison of our method with existing UDA, SSL, and SSDA methods on \textbf{Synthia $\to$ Cityscapes} benchmark. We report 16-class mIoU scores on the Cityscapes validation set and follow the same settings as in \autoref{tab:comparison-GTA}. DACS++\textsuperscript{\textdagger} represents implementation of DACS \cite{tranheden2021dacs} from UDA to SSDA. }
\label{tab:comparison-synthia}
\resizebox{\columnwidth}{!}{%
\begin{tabular}{@{}cccccccc@{}}
\toprule
 &  & \multicolumn{6}{c}{\textbf{Labeled Target Samples}} \\ \cmidrule(l){3-8} 
\multirow{-2}{*}{\textbf{Type}} & \multirow{-2}{*}{\textbf{Methods}} & \textbf{0} & \textbf{100} & \textbf{200} & \textbf{500} & \textbf{1000} & \textbf{2975} \\ \midrule
Supervised & DeepLabV2 \cite{chen2017deeplab}& - & 53.0 & 58.9 & 61.0 & 67.5 & 70.8 \\ \midrule
 & DaCS \cite{tranheden2021dacs} & 54.8 & - & - & - & - & - \\
 & PRoDA \cite{zhang2021prototypical}& {\color[HTML]{3531FF} 62.0} & - & - & - & - & - \\
 & DaFormer \cite{hoyer2022daformer}& { 58.8} & - & - & - & - & - \\
\multirow{-4}{*}{UDA} & DIGA \cite{shen2023diga}& {\color[HTML]{FE0000} 67.9} & - & - & - & - & - \\ \midrule
 & CowMix \cite{french2019semi}& - & 61.3 & \color[HTML]{FFFFFF}66.7 & \color[HTML]{FFFFFF}71.1 & \color[HTML]{FFFFFF}73.0 & - \\
 & ClassMix \cite{olsson2021classmix}& - & {\color[HTML]{3531FF} 61.4} & \color[HTML]{FFFFFF}{\color[HTML]{3531FF} 67.6} & \color[HTML]{FFFFFF}{\color[HTML]{3531FF} 72.3} & \color[HTML]{FFFFFF}{\color[HTML]{3531FF} 73.1} & \multicolumn{1}{l}{} \\
 & DMT \cite{feng2022dmt}& - & 59.7 & \color[HTML]{FFFFFF}64.3 & \color[HTML]{FFFFFF}68.9 & \color[HTML]{FFFFFF}70.5 & - \\
\multirow{-4}{*}{SSL} & DusPerb \cite{yang2023revisiting}& - & {\color[HTML]{FE0000} 68.4} & \color[HTML]{FFFFFF}{\color[HTML]{FE0000} 71.4} & \color[HTML]{FFFFFF}{\color[HTML]{FE0000} 74.2} & \color[HTML]{FFFFFF}{\color[HTML]{FE0000} 76.1} & - \\ \midrule
 & ASS \cite{xie2022towards}& - & 62.1 & 64.8 & 69.8 & 73.0 & 77.1 \\
 & ComplexMix \cite{chen2021complexmix}& - & 70.6 & \color[HTML]{FFFFFF}71.8 & \color[HTML]{FFFFFF}72.6 & \color[HTML]{FFFFFF}74.0 & 75.6 \\
 & DACS++\textsuperscript{\textdagger} \cite{tranheden2021dacs}& - & 64.9 & 67.7 & 71.3 & 72.8 & 74.4 \\
 & DLDM \cite{chen2021semidual}& - & 68.4 & \color[HTML]{FFFFFF}69.8 & \color[HTML]{FFFFFF}71.7 & \color[HTML]{FFFFFF}74.2 & \color[HTML]{FFFFFF}77.2 \\
 & SSDDA \cite{chen2022semi}& - & 70.6 & 71.8 & 72.6 & 74.0 & - \\
 & ALFSA \cite{wen2024semi} & - & 68.9 & \color[HTML]{FFFFFF}{\color[HTML]{3531FF} 73.5} & \color[HTML]{FFFFFF}{\color[HTML]{FE0000} 77.5} & \color[HTML]{FFFFFF}{\color[HTML]{FE0000} 79.0} & \color[HTML]{FFFFFF}{\color[HTML]{FE0000} 79.9} \\
 & SLA \cite{yu2023semi}& - & 63.7 & \color[HTML]{FFFFFF}66.1 & \color[HTML]{FFFFFF}{ 71.9} & \color[HTML]{FFFFFF}{ 74.4} & \color[HTML]{FFFFFF}{ 77.8} \\
 & S-Depth \cite{hoyer2023improving}& - & {\color[HTML]{3531FF} 72.4} & {\color[HTML]{3531FF} 73.5} & 75.4 & 76.3 & 77.1 \\
 & IIDM \cite{fu2024iidminterintradomainmixing}& - & {\color[HTML]{FE0000} 74.2} & {\color[HTML]{FE0000} 76.4} & {\color[HTML]{3531FF} 77.0} & {\color[HTML]{3531FF} 78.8} & {\color[HTML]{3531FF} 79.2} \\ \cmidrule(l){2-8} 
 & \textbf{Ours w/o DyCE} & \textbf{69.5} & \textbf{74.9} & \textbf{76.8} & \textbf{77.7} & \textbf{79.2} & \textbf{79.6} \\ \cmidrule(l){2-8} 
\multirow{-11}{*}{SSDA} & \textbf{Ours w/ DyCE} & \textbf{70.2} & \textbf{76.9} & \textbf{77.2} & \textbf{78.6} & \textbf{79.7} & \textbf{80.5} \\ \bottomrule
\end{tabular}%
}
\end{table}

\begin{table}[tbp]
\centering
\caption{Ablation experiments using three different SSDA settings on \textbf{GTA5$\to$Cityscapes} and \textbf{Synthia$\to$Cityscapes} to identify the contribution of individual components: Consistency Training (CT), Dynamic Cross-Entropy loss (DyCE), Vision-Language Pre-training (VLP), and DenseLanguage Guidance (DLG). }
\label{tab:ablation}
\resizebox{\columnwidth}{!}{%
\begin{tabular}{@{}
c 
c 
c 
c 
c 
c 
c 
c 
c 
c 
c 
c @{}}
\toprule
\multicolumn{4}{c}{\textbf{Components}} & \multicolumn{4}{c}{\textbf{GTA5 $\to$ Cityscapes}} & \multicolumn{4}{c}{\textbf{Synthia $\to$ Cityscapes}} \\ \midrule
\textbf{CT} & \textbf{DyCE} & \textbf{VLP} & \textbf{DLG} & \textbf{100} & \textbf{200} & \textbf{500} & \textbf{1000} & \textbf{100} & \textbf{200} & \textbf{500} & \textbf{1000} \\ \midrule
$\checkmark$ & - & - & - & 54.5 & 58.2 & 62.3 & 64.6 & 60.2 & 65.3 & 71.5 & 72.0 \\
$\checkmark$ & $\checkmark$ & - & - & 63.3 & 64.5 & 65.9 & 68.2 & 68.7 & 70.1 & 72.2 & 74.7 \\
$\checkmark$ & - & $\checkmark$ & - & 65.6 & 66.8 & 69.1 & 69.9 & 71.9 & 73.0 & 74.7 & 75.9 \\
$\checkmark$ & - & $\checkmark$ & $\checkmark$ & 70.3 & 71.6 & 72.1 & 73.9 & 74.9 & 76.8 & 77.7 & 79.2 \\
\textbf{$\checkmark$} & \textbf{$\checkmark$} & \textbf{$\checkmark$} & \textbf{$\checkmark$} & \textbf{71.1} & \textbf{72.5} & \textbf{72.9} & \textbf{74.8} & \textbf{76.9} & \textbf{77.2} & \textbf{78.6} & \textbf{79.7} \\ \bottomrule
\end{tabular}%
}
\vspace{-4mm}
\end{table}

\begin{table*}[tbp]
\centering
\caption{Class-wise performance evaluation of our proposed method (with and without the proposed class-balancing DyCE loss), and comparison with the existing class-balanced UDA and SSDA methods. We report 19-class and 16-class mIoU scores on the \textbf{GTA5 $\to$ Cityscapes} and \textbf{Synthia $\to$ Cityscapes} settings, respectively with 100 labeled target samples. The segmentation performance of tailed classes significantly improves by incorporating our DyCe loss in both settings. Our results are \textbf{highlighted} whereas the previous-best and second-best results are marked in \textcolor{red}{red} and \textcolor{blue}{blue}. Please refer to \textbf{supplementary file} for detailed class distribution and improvement analysis. }
\label{tab:class-wise-comparison}
\resizebox{2\columnwidth}{!}{%
\begin{tabular}{@{}ccccccccccccccccccccccc@{}}
\toprule

{ } & { } & { } & \multicolumn{20}{c}{{ \textbf{GTA5 $\to$ Cityscapes}}} \\ \cmidrule(l){4-23} 

\multirow{-2}{*}{{ \textbf{Type}}} & \multirow{-2}{*}{{ \textbf{Methods}}} & \multirow{-2}{*}{{ \textbf{\begin{tabular}[c]{@{}c@{}}Target\\ Labels\end{tabular}}}} & { \textbf{Road}} & { \textbf{Sidewalk}} & { \textbf{Building}} & { \textbf{Walls}} & { \textbf{Fence}} & { \textbf{Pole}} & { \textbf{T-Light}} & { \textbf{T-sign}} & { \textbf{Veg}} & { \textbf{Terrain}} & { \textbf{Sky}} & { \textbf{Person}} & { \textbf{Rider}} & { \textbf{Car}} & { \textbf{Truck}} & { \textbf{Bus}} & { \textbf{Train}} & { \textbf{Motor}} & { \textbf{Bike}} & { \textbf{mIoU}} \\ \midrule

{ } & { UniMatch} \cite{yang2023revisiting}& { } & { 97.2} & { 79.3} & { 90.6} & { 36.5} & { 52.1} & { 56.7} & { 64.2} & { 72.1} & { 91.1} & { 59.0} & { 93.6} & { 77.5} & { 53.5} & { 93.4} & { 73.8} & { 79.8} & { 67.8} & { 49.6} & { 71.2} & { 71.5} \\

\multirow{-2}{*}{{ \textbf{Supervised}}} & { U2PL} \cite{wang2022semiu2pl}& \multirow{-2}{*}{{ 2975}} & { 97.6} & { 82.1} & { 90.7} & { 39.3} & { 53.4} & { 58.0} & { 64.2} & { 74.0} & { 90.9} & { 60.2} & { 93.7} & { 77.0} & { 49.8} & { 93.7} & { 64.8} & { 77.9} & { 47.9} & { 51.2} & { 73.6} & { 70.5} \\ \midrule

{ } & { CADA} \cite{yang2021context} & { } & { 91.3} & { 46.0} & { 84.5} & { 34.4} & { 29.7} & { 32.6} & { 35.8} & { 36.4} & { 84.5} & { 43.2} & { 83.0} & { 60.0} & { 32.2} & { 83.2} & { 35.0} & { 46.7} & { 0.0} & { 33.7} & { 42.2} & { 49.2} \\

{ } & { IAST \cite{mei2020instance}} & { } & {\color[HTML]{FE0000} 93.8} & {\color[HTML]{3531FF} 57.8} & { 85.1} & { 39.5} & { 26.7} & { 26.2} & { 43.1} & { 34.7} & { 84.9} & { 32.9} & {\color[HTML]{3531FF} 88.0} & { 62.6} & { 29.0} & { 87.3} & { 39.2} & { 49.6} & {\color[HTML]{3531FF} 23.2} & { 34.7} & { 39.6} & { 51.5} \\

{ } & { DACS} \cite{tranheden2021dacs}& { } & { 89.9} & { 39.7} & {\color[HTML]{FE0000} 87.9} & { 30.7} & {\color[HTML]{3531FF} 39.5} & { 38.5} & { 46.4} & { 52.8} & {\color[HTML]{3531FF} 88.0} & {\color[HTML]{3531FF} 44.0} & {\color[HTML]{FE0000} 88.8} & { 67.2} & { 35.8} & { 84.5} & {\color[HTML]{3531FF} 45.7} & { 50.2} & { 0.0} & { 27.3} & { 34.0} & { 52.1} \\

{ } & { Shallow} \cite{cardace2022shallow}& { } & { 91.9} & { 48.9} & {\color[HTML]{3531FF} 86.0} & { 38.6} & { 28.6} & { 34.8} & { 45.6} & { 43.0} & { 86.2} & { 42.4} & { 87.6} & { 65.6} & {\color[HTML]{3531FF} 38.6} & { 86.8} & { 38.4} & { 48.2} & { 0.0} & { 46.5} & {\color[HTML]{FE0000} 59.2} & { 53.5} \\

{ } & { ProDA+distill} \cite{li2022class}& { } & { 87.8} & { 56.0} & { 79.7} & {\color[HTML]{FE0000} 46.3} & {\color[HTML]{FE0000} 44.8} & {\color[HTML]{3531FF} 45.6} & {\color[HTML]{3531FF} 53.5} & {\color[HTML]{3531FF} 53.5} & {\color[HTML]{FE0000} 88.6} & {\color[HTML]{FE0000} 45.2} & { 82.1} & {\color[HTML]{3531FF} 70.7} & {\color[HTML]{FE0000} 39.2} & {\color[HTML]{3531FF} 88.8} & { 45.5} & {\color[HTML]{3531FF} 59.4} & { 1.0} & {\color[HTML]{3531FF} 48.9} & {\color[HTML]{3531FF} 56.4} & {\color[HTML]{3531FF} 57.5} \\

\multirow{-6}{*}{{ \textbf{UDA}}} & { CPSL+distill} \cite{li2022class}& \multirow{-6}{*}{{ 0}} & {\color[HTML]{3531FF} 92.3} & {\color[HTML]{FE0000} 59.9} & { 84.9} & {\color[HTML]{3531FF} 45.7} & { 29.7} & {\color[HTML]{FE0000} 52.8} & {\color[HTML]{FE0000} 61.5} & {\color[HTML]{FE0000} 59.5} & { 87.9} & { 41.5} & { 85.0} & {\color[HTML]{FE0000} 73.0} & { 35.5} & {\color[HTML]{FE0000} 90.4} & {\color[HTML]{FE0000} 48.7} & {\color[HTML]{FE0000} 73.9} & {\color[HTML]{FE0000} 26.3} & {\color[HTML]{FE0000} 53.8} & { 53.9} & {\color[HTML]{FE0000} 60.8} \\ \midrule

{ } & { ALFSA} \cite{wen2024semi}& { } & {\color[HTML]{FE0000} 95.9} & {\color[HTML]{FE0000} 71.5} & {\color[HTML]{FE0000} 87.4} & {\color[HTML]{FE0000} 39.9} & {\color[HTML]{FE0000} 39.0} & {\color[HTML]{FE0000} 44.6} & {\color[HTML]{FE0000} 52.6} & {\color[HTML]{3531FF} 60.4} & {\color[HTML]{3531FF} 89.1} & {\color[HTML]{3531FF} 50.7} & {\color[HTML]{3531FF} 91.3} & {\color[HTML]{FE0000} 73.1} & {\color[HTML]{FE0000} 48.3} & {\color[HTML]{FE0000} 91.3} & {\color[HTML]{3531FF} 55.3} & {\color[HTML]{FE0000} 63.7} & {\color[HTML]{3531FF} 26.3} & {\color[HTML]{FE0000} 55.8} & {\color[HTML]{FE0000} 68.7} & {\color[HTML]{FE0000} 63.4} \\

{ } & { SS-ADA+UniMatch} \cite{yan2024ss}& { } & {\color[HTML]{3531FF} 96.4} & {\color[HTML]{3531FF} 75.0} & {\color[HTML]{3531FF} 89.2} & {\color[HTML]{3531FF} 43.7} & {\color[HTML]{3531FF} 45.1} & {\color[HTML]{3531FF} 53.3} & {\color[HTML]{3531FF} 58.2} & {\color[HTML]{FE0000} 68.8} & {\color[HTML]{FE0000} 90.7} & {\color[HTML]{FE0000} 55.4} & {\color[HTML]{FE0000} 93.8} & {\color[HTML]{3531FF} 75.8} & {\color[HTML]{3531FF} 49.7} & {\color[HTML]{3531FF} 91.6} & { 54.6} & {\color[HTML]{3531FF} 67.4} & {\color[HTML]{FE0000} 43.6} & { 47.2} & {\color[HTML]{3531FF} 69.4} & {\color[HTML]{3531FF} 66.8} \\

{ } & SS-ADA+U2PL \cite{yan2024ss}& { } & {\color[HTML]{FE0000} 96.5} & {\color[HTML]{FE0000} 75.5} & {\color[HTML]{FE0000} 89.7} & {\color[HTML]{FE0000} 47.1} & {\color[HTML]{FE0000} 47.7} & {\color[HTML]{FE0000} 55.3} & {\color[HTML]{FE0000} 60.6} & {\color[HTML]{3531FF} 68.1} & {\color[HTML]{3531FF} 90.6} & {\color[HTML]{3531FF} 55.3} & {\color[HTML]{3531FF} 92.1} & {\color[HTML]{FE0000} 77.4} & {\color[HTML]{FE0000} 52.5} & {\color[HTML]{FE0000} 92.5} & {\color[HTML]{FE0000} 67.1} & {\color[HTML]{FE0000} 67.8} & {\color[HTML]{3531FF} 41.2} & {\color[HTML]{3531FF} 49.9} & {\color[HTML]{FE0000} 70.8} & {\color[HTML]{FE0000} 68.3} \\

{ } & { \textbf{Ours w/o DyCE}} & { } & { \textbf{97.3}} & { \textbf{80.2}} & { \textbf{89.5}} & { \textbf{50.2}} & { \textbf{49.2}} & { \textbf{58.3}} & { \textbf{62.3}} & { \textbf{69.3}} & { \textbf{90.6}} & { \textbf{57.7}} & { \textbf{93.2}} & { \textbf{78.6}} & { \textbf{53.9}} & { \textbf{92.9}} & { \textbf{68.4}} & { \textbf{67.9}} & { \textbf{47.9}} & { \textbf{56.5}} & { \textbf{70.9}} & { \textbf{70.3}} \\

\multirow{-5}{*}{{ \textbf{SSDA}}} & { \textbf{Ours w/ DyCE}} & \multirow{-5}{*}{{ 100}} & { \textbf{98.1}} & { \textbf{80.9}} & { \textbf{90.3}} & { \textbf{51.9}} & { \textbf{51.5}} & { \textbf{60.2}} & { \textbf{62.5}} & { \textbf{71.1}} & { \textbf{90.9}} & { \textbf{59.5}} & { \textbf{93.2}} & { \textbf{79.9}} & { \textbf{56.8}} & { \textbf{93.2}} & { \textbf{71.6}} & { \textbf{68.1}} & { \textbf{49.1}} & { \textbf{58.6}} & { \textbf{71.4}} & { \textbf{71.1}} \\ \midrule

{ } & { \textbf{}} & { \textbf{}} & \multicolumn{20}{c}{{ \textbf{Synthia $\to$ Cityscapes}}} \\ \midrule

{ } & { UniMatch} \cite{yang2023revisiting}& { } & { 97.5} & { 82.1} & { 91.2} & { 52.4} & { 53.0} & { 60.7} & { 66.3} & { 75.3} & { 92.3} & { -} & { 94.1} & { 79.9} & { 57.5} & { 94.4} & { -} & { 82.1} & { -} & { 57.9} & { 74.5} & { 75.7} \\

\multirow{-2}{*}{{ \textbf{Supervised}}} & { U2PL} \cite{wang2022semiu2pl}& \multirow{-2}{*}{{ 2975}} & { 97.5} & { 81.7} & { 90.0} & { 36.9} & { 50.9} & { 56.8} & { 59.9} & { 71.7} & { 91.6} & { -} & { 93.1} & { 76.5} & { 43.5} & { 93.6} & { -} & { 75.4} & { -} & { 45.2} & { 72.1} & { 71.0} \\ \midrule

{ } & { FADA} {\cite{wang2020classes}} & { } & { 84.5} & { 40.1} & { 83.1} & { 4.8} & { 0.0} & { 34.3} & { 20.1} & { 27.2} & { 84.8} & { -} & { 84.0} & { 53.5} & { 22.6} & { 85.4} & { -} & { 43.7} & { -} & { 26.8} & { 27.8} & { 45.2} \\

{ } & { IAST} {\cite{mei2020instance}} & { } & { 81.9} & { 41.5} & { 83.3} & { 17.7} & {\color[HTML]{FE0000} 4.6} & { 32.3} & { 30.9} & { 28.8} & { 83.4} & { -} & { 85.0} & { 65.5} & { 30.8} & { 86.5} & { -} & { 38.2} & { -} & { 33.1} & { 52.7} & { 49.8} \\

{ } & { DACS {}} \cite{tranheden2021dacs}& { } & { 80.6} & { 25.1} & { 81.9} & { 21.5} & {\color[HTML]{3531FF} 2.6} & { 37.2} & { 22.7} & { 24.0} & { 83.7} & { -} & {\color[HTML]{FE0000} 90.8} & { 67.6} & {\color[HTML]{FE0000} 38.3} & { 82.9} & { -} & { 38.9} & { -} & { 28.5} & {\color[HTML]{3531FF} 47.6} & { 48.3} \\

{ } & { Shallow} \cite{cardace2022shallow}& { } & {\color[HTML]{FE0000} 90.4} & {\color[HTML]{FE0000} 51.1} & { 83.4} & { 3.0} & { 0.0} & { 32.3} & { 25.3} & { 31.0} & { 84.8} & { -} & { 85.5} & { 59.3} & { 30.1} & { 82.6} & { -} & {\color[HTML]{FE0000} 53.2} & { -} & { 17.5} & { 45.6} & { 48.4} \\

{ } & { ProDA+distill} \cite{li2022class}& { } & {\color[HTML]{3531FF} 87.8} & {\color[HTML]{3531FF} 45.7} & {\color[HTML]{3531FF} 84.6} & {\color[HTML]{FE0000} 37.1} & { 0.6} & {\color[HTML]{3531FF} 44.0} & {\color[HTML]{3531FF} 54.6} & {\color[HTML]{3531FF} 37.0} & {\color[HTML]{FE0000} 88.1} & { -} & { 84.4} & {\color[HTML]{3531FF} 74.2} & { 24.3} & {\color[HTML]{3531FF} 88.2} & { -} & {\color[HTML]{3531FF} 51.1} & { -} & {\color[HTML]{3531FF} 40.5} & { 45.6} & {\color[HTML]{3531FF} 55.5} \\

\multirow{-6}{*}{{ \textbf{UDA}}} & { CPSL+distill} \cite{li2022class}& \multirow{-6}{*}{{ 0}} & { 87.2} & { 43.9} & {\color[HTML]{FE0000} 85.5} & {\color[HTML]{3531FF} 33.6} & { 0.3} & {\color[HTML]{FE0000} 47.7} & {\color[HTML]{FE0000} 57.4} & {\color[HTML]{FE0000} 37.2} & {\color[HTML]{3531FF} 87.8} & { -} & {\color[HTML]{3531FF} 88.5} & {\color[HTML]{FE0000} 79.0} & {\color[HTML]{3531FF} 32.0} & {\color[HTML]{FE0000} 90.6} & { -} & { 49.4} & { -} & {\color[HTML]{FE0000} 50.8} & {\color[HTML]{FE0000} 59.8} & {\color[HTML]{FE0000} 57.9} \\ \midrule

{ } & { SS-ADA+U2PL} \cite{yan2024ss}& { } & {\color[HTML]{3531FF} 91.0} & {\color[HTML]{3531FF} 62.0} & {\color[HTML]{3531FF} 86.7} & {\color[HTML]{3531FF} 38.9} & {\color[HTML]{3531FF} 33.4} & {\color[HTML]{3531FF} 53.6} & {\color[HTML]{FE0000} 58.9} & {\color[HTML]{3531FF} 69.0} & {\color[HTML]{3531FF} 91.0} & { -} & {\color[HTML]{3531FF} 92.5} & {\color[HTML]{3531FF} 73.9} & {\color[HTML]{3531FF} 44.6} & {\color[HTML]{3531FF} 92.3} & { -} & {\color[HTML]{FE0000} 69.3} & { -} & {\color[HTML]{3531FF} 37.3} & {\color[HTML]{3531FF} 67.2} & {\color[HTML]{3531FF} 66.4} \\

{ } & { SS-ADA+UniMatch} \cite{yan2024ss}& { } & {\color[HTML]{FE0000} 97.1} & {\color[HTML]{FE0000} 79.4} & {\color[HTML]{FE0000} 90.2} & {\color[HTML]{FE0000} 49.8} & {\color[HTML]{FE0000} 49.8} & {\color[HTML]{FE0000} 56.9} & {\color[HTML]{3531FF} 58.2} & {\color[HTML]{FE0000} 72.2} & {\color[HTML]{FE0000} 91.6} & { -} & {\color[HTML]{FE0000} 93.4} & {\color[HTML]{FE0000} 78.1} & {\color[HTML]{FE0000} 53.3} & {\color[HTML]{FE0000} 92.8} & { -} & {\color[HTML]{3531FF} 69.1} & { -} & {\color[HTML]{FE0000} 48.4} & {\color[HTML]{FE0000} 72.1} & {\color[HTML]{FE0000} 72.0} \\

{ } & { \textbf{Ours w/o DyCE}} & { } & { \textbf{98.5}} & { \textbf{78.9}} & { \textbf{91.6}} & { \textbf{52.7}} & { \textbf{52.8}} & { \textbf{62.3}} & { \textbf{63.9}} & { \textbf{74.3}} & { \textbf{91.5}} & { \textbf{-}} & { \textbf{94.4}} & { \textbf{79.7}} & { \textbf{56.9}} & { \textbf{93.1}} & { \textbf{-}} & { \textbf{76.6}} & { \textbf{-}} & { \textbf{55.5}} & { \textbf{74.9}} & { \textbf{74.9}} \\

\multirow{-4}{*}{{ \textbf{SSDA}}} & { \textbf{Ours w/ DyCE}} & \multirow{-4}{*}{{ 100}} & { \textbf{98.9}} & { \textbf{80.9}} & { \textbf{92.2}} & { \textbf{57.6}} & { \textbf{56.2}} & { \textbf{63.8}} & { \textbf{67.1}} & { \textbf{76.7}} & { \textbf{91.9}} & { \textbf{-}} & { \textbf{95.9}} & { \textbf{80.6}} & { \textbf{59.9}} & { \textbf{93.8}} & { \textbf{-}} & { \textbf{78.9}} & { \textbf{-}} & { \textbf{59.8}} & { \textbf{76.6}} & { \textbf{76.9}} \\ \bottomrule
\end{tabular}%
}
\vspace{-3mm}
\end{table*}

Our proposed SemiDAViL framework demonstrates significant gains on both GTA5$\to$Cityscapes and Synthia$\to$Cityscapes benchmarks (\autoref{tab:comparison-GTA}, \autoref{tab:comparison-synthia}). We compare against state-of-the-art UDA, SSL, and SSDA techniques, highlighting its robustness with varying levels of target annotations.

In the \textbf{GTA5$\to$Cityscapes} scenario, \textbf{(A)} using only 100 labeled target samples, SemiDAViL achieves 71.1\% mIoU, outperforming the previous best, IIDM \cite{fu2024iidminterintradomainmixing}, by 1.6\%. The advantage grows with 200 labeled samples, where we attain 72.5\% mIoU, showcasing our framework's strength in leveraging limited annotations through language-guided features; \textbf{(B)} in the fully unsupervised setting, our method achieves 67.7\% mIoU, a 5\% improvement over the previous best, DIGA \cite{shen2023diga}, owing to our language-guided joint embedding, which provides more robust semantic alignment than divergence-based methods like DaFormer \cite{hoyer2022daformer}; and \textbf{(C)} with 2975 labeled samples, our model reaches 75.2\% mIoU, surpassing the previous best, IIDM by 1.9\%, attributed to the effective handling of class imbalance.

For \textbf{Synthia$\to$Cityscapes}, similar gains are observed against the previous best methods like IIDM \cite{fu2024iidminterintradomainmixing} and ALFSA \cite{wen2024semi}, demonstrating the strength of our dense language guidance and adaptive DyCE loss. Competing methods, such as DACS++ \cite{tranheden2021dacs}, attempt to address class imbalance via pseudo-label refinement, but their reliance on multi-stage training and hyperparameter tuning limits scalability. Unlike these, our end-to-end solution with DyCE loss adaptively re-weights based on class distribution, yielding balanced learning without the need for manual adjustments. 

Overall, our framework consistently outperforms across all benchmarks and supervision levels, demonstrating its robustness and scalability. Detailed class-wise analysis is provided in \autoref{subsection:loss-validation}.

\begin{table*}[tbp]
\centering
\caption{Comparative performance improvement by incorporating our proposed DyCE loss to address \textbf{class imbalance} in 20\% labeled Synapse dataset in SSL setting. Please refer to \textbf{supplementary file} for detailed class-distribution and improvement analysis.}
\label{tab:medical-comparison}
\resizebox{2.2\columnwidth}{!}{%
\begin{tabular}{@{}ccccccccccccccccc@{}}
\toprule
 &  & \multicolumn{2}{c}{\textbf{Average}} & \multicolumn{13}{c}{\textbf{Average DSC$\uparrow$ of Each Class}} \\ \cmidrule(l){3-4} 
\multirow{-2}{*}{\textbf{Type}} & \multirow{-2}{*}{\textbf{Methods}} & \textbf{DSC} $\uparrow$& \textbf{ASD} $\downarrow$& \textbf{Sp} & \textbf{RK} & \textbf{LK} & \textbf{Ga} & \textbf{Es} & \textbf{Li} & \textbf{St} & \textbf{Ao} & \textbf{IVC} & \textbf{PSV} & \textbf{PA} & \textbf{RAG} & \textbf{LAG} \\ \midrule
Supervised & V-Net \cite{milletari2016v}& 62.1 & 10.3 & 84.6 & 77.2 & 73.8 & 73.3 & 38.2 & 94.6 & 68.4 & 72.1 & 71.2 & 58.2 & 48.5 & 17.9 & 29.0 \\ \midrule
 & { UA-MT \cite{yu2019uncertainty}} & { 20.3} & { 71.7} & { 48.2} & { 31.7} & { 22.2} & { 0.0} & { 0.0} & { 81.2} & { 29.1} & { 23.3} & { 27.5} & { 0.0} & { 0.0} & { 0.0} & { 0.0} \\
 & { URPC \cite{luo2022semi}} & { 25.7} & { 72.7} & { 60.7} & { 38.2} & { 56.8} & { 0.0} & { 0.0} & { 85.3} & { 33.9} & { 33.1} & { 14.8} & { 0.0} & { 5.1} & { 0.0} & { 0.0} \\
 & CPS \cite{chen2021semicps} & 33.6 & 41.2 & 62.8 & 55.2 & 45.4 & 35.9 & 0.0 & 91.1 & 31.3 & 41.9 & 49.2 & 8.8 & 14.5 & 0.0 & 0.0 \\
 & SS-Net \cite{wu2022exploring} & 35.1 & 50.8 & 62.7 & 67.9 & 60.9 & 34.3 & 0.0 & 89.9 & 20.9 & 61.7 & 44.8 & 0.0 & 8.7 & 4.2 & 0.0 \\
 & DST \cite{chen2022debiased} & 34.5 & 37.7 & 57.7 & 57.2 & 46.4 & 43.7 & 0.0 & 89.0 & 33.9 & 43.3 & 46.9 & 9.0 & 21.0 & 0.0 & 0.0 \\
\multirow{-6}{*}{\begin{tabular}[c]{@{}c@{}}General\\ SSL\end{tabular}} & { DePL \cite{wang2022debiased}} & { 36.3} & { 36.0} & { 62.8} & { 61.0} & { 48.2} & { 54.8} & { 0.0} & { 90.2} & { 36.0} & { 42.5} & { 48.2} & { 10.7} & { 17.0} & { 0.0} & { 0.0} \\ \midrule
 & Adsh \cite{guo2022class} & 35.3 & 39.6 & 55.1 & 59.6 & 45.8 & 52.2 & 0.0 & 89.4 & 32.8 & 47.6 & 53.0 & 8.9 & 14.4 & 0.0 & 0.0 \\
 & CReST \cite{wei2021crest} & 38.3 & 22.9 & 62.1 & 64.7 & 53.8 & 43.8 & 8.1 & 85.9 & 27.2 & 54.4 & 47.7 & 14.4 & 13.0 & 18.7 & 4.6 \\
 & SimiS \cite{chen2022embarrassingly} & 40.1 & 33.0 & 62.3 & 69.4 & 50.7 & 61.4 & 0.0 & 87.0 & 33.0 & 59.0 & 57.2 & 29.2 & 11.8 & 0.0 & 0.0 \\
 & ACISSMIS \cite{basak2022addressing} & 33.2 & 43.8 & 57.4 & 53.8 & 48.5 & 46.9 & 0.0 & 87.8 & 28.7 & 42.3 & 45.4 & 6.3 & 15.0 & 0.0 & 0.0 \\
 & CLD \cite{lin2022calibrating} & 41.1 & 32.2 & 62.0 & 66.0 & 59.3 & 61.5 & 0.0 & 89.0 & 31.7 & 62.8 & 49.4 & 28.6 & 18.5 & 0.0 & 5.0 \\
 & {\ DHC \cite{wang2023dhc}} & {\ 48.6} & {\ 10.7} & {\ 62.8} & {\ 69.5} & {\ 59.2} & {\ 66.0} & {\ 13.2} & {\ 85.2} & {\ 36.9} & {\ 67.9} & {\ 61.5} & {\ 37.0} & {\ 30.9} & {\ 31.4} & {\ 10.6} \\
\multirow{-7}{*}{\begin{tabular}[c]{@{}c@{}}Class-balanced\\ SSL\end{tabular}} & { A\&D} \cite{wang2024towards}& { 60.9} & { 2.5} & { 85.2} & { 66.9} & { 67.0} & { 52.7} & { 62.9} & { 89.6} & { 52.1} & { 83.0} & { 74.9} & { 41.8} & { 43.4} & { 44.8} & { 27.2} \\ \midrule
 & { \textbf{UA-MT\cite{yu2019uncertainty}+DyCE}} & { \textbf{48.1}}(\textcolor{black}{+27.9}) & { \textbf{10.9}}(\textcolor{black}{-60.8}) & { \textbf{55.6}}(\textcolor{black}{+7.4}) & { \textbf{52.3}}(\textcolor{black}{+20.6}) & { \textbf{53.4}}(\textcolor{black}{+31.2}) & { \textbf{49.5}}(\textcolor{black}{+49.5}) & { \textbf{51.5}}(\textcolor{black}{+51.5}) & { \textbf{88.7}}(\textcolor{black}{+7.5}) & { \textbf{46.0}}(\textcolor{black}{+16.9}) & { \textbf{48.7}}(\textcolor{black}{+25.4}) & { \textbf{47.3}}(\textcolor{black}{+19.8}) & { \textbf{35.6}}(\textcolor{black}{+35.6}) & { \textbf{33.4}}(\textcolor{black}{+33.4}) & { \textbf{39.1}}(\textcolor{black}{+39.1}) & { \textbf{24.4}}(\textcolor{black}{+24.4}) \\
 & { \textbf{URPC\cite{luo2022semi}+DyCE}} & { \textbf{48.9}}(\textcolor{black}{+23.2}) & { \textbf{10.1}}(\textcolor{black}{-62.6}) & { \textbf{63.2}}(\textcolor{black}{+2.5}) & { \textbf{59.9}}(\textcolor{black}{+21.7}) & { \textbf{59.4}}(\textcolor{black}{+2.6}) & { \textbf{52.5}}(\textcolor{black}{+52.5}) & { \textbf{52.2}}(\textcolor{black}{+52.2}) & { \textbf{89.3}}(\textcolor{black}{+4.0}) & { \textbf{45.6}}(\textcolor{black}{+11.7}) & { \textbf{50.6}}(\textcolor{black}{+17.5}) & { \textbf{50.3}}(\textcolor{black}{+35.5}) & { \textbf{30.1}}(\textcolor{black}{+30.1}) & { \textbf{30.4}}(\textcolor{black}{+25.3}) & { \textbf{26.7}}(\textcolor{black}{+26.7}) & { \textbf{25.1}}(\textcolor{black}{+25.1}) \\
 & {\ \textbf{DePL\cite{wang2022debiased}+DyCE}} & {\ \textbf{51.8}}(\textcolor{black}{+15.6}) & {\ \textbf{9.2}}(\textcolor{black}{-26.1}) & {\ \textbf{65.9}}(\textcolor{black}{+3.1}) & {\ \textbf{64.2}}(\textcolor{black}{+3.2}) & {\ \textbf{65.1}}(\textcolor{black}{+16.9}) & {\ \textbf{59.5}}(\textcolor{black}{+4.7}) & {\ \textbf{58.2}}(\textcolor{black}{+58.2}) & {\ \textbf{90.5}}(\textcolor{black}{+0.3}) & {\ \textbf{41.7}}(\textcolor{black}{+5.7}) & {\ \textbf{54.9}}(\textcolor{black}{+12.4}) & {\ \textbf{55.6}}(\textcolor{black}{+7.4}) & {\ \textbf{32.2}}(\textcolor{black}{+21.5}) & {\ \textbf{35.8}}(\textcolor{black}{+18.8}) & {\ \textbf{26.9}}(\textcolor{black}{+26.9}) & {\ \textbf{23.3}}(\textcolor{black}{+23.3}) \\
 & {\ \textbf{DHC\cite{wang2023dhc}+DyCE}} & {\ \textbf{57.9}}(\textcolor{black}{+9.3}) & {\ \textbf{6.4}}(\textcolor{black}{-4.3}) & {\ \textbf{77.5}}(\textcolor{black}{+14.7}) & {\ \textbf{72.5}}(\textcolor{black}{+3.0}) & {\ \textbf{64.3}}(\textcolor{black}{+5.1}) & {\ \textbf{70.3}}(\textcolor{black}{+4.3}) & {\ \textbf{53.2}}(\textcolor{black}{+40.0}) & {\ \textbf{88.7}}(\textcolor{black}{+3.5}) & {\ \textbf{44.8}}(\textcolor{black}{+7.9}) & {\ \textbf{79.1}}(\textcolor{black}{+11.2}) & {\ \textbf{67.8}}(\textcolor{black}{+6.3}) & {\ \textbf{42.0}}(\textcolor{black}{+5.0}) & {\ \textbf{38.2}}(\textcolor{black}{+7.3}) & {\ \textbf{34.8}}(\textcolor{black}{+3.4}) & {\ \textbf{20.1}}(\textcolor{black}{+9.5}) \\
\multirow{-5}{*}{\begin{tabular}[c]{@{}c@{}}\textbf{SSL+Our}\\\textbf{DyCE loss}\end{tabular}} & { \textbf{A\&D\cite{wang2024towards}+DyCE}} & { \textbf{65.5}}(\textcolor{black}{+4.6}) & { \textbf{5.8}}(\textcolor{black}{-0.7}) & { \textbf{87.3}}(\textcolor{black}{+2.1}) & { \textbf{78.5}}(\textcolor{black}{+11.6}) & { \textbf{72.4}}(\textcolor{black}{+5.4}) & { \textbf{65.1}}(\textcolor{black}{+12.4}) & { \textbf{64.7}}(\textcolor{black}{+1.8}) & { \textbf{90.9}}(\textcolor{black}{+1.3}) & { \textbf{55.3}}(\textcolor{black}{+3.2}) & { \textbf{84.1}}(\textcolor{black}{+1.1}) & { \textbf{77.5}}(\textcolor{black}{+2.6}) & { \textbf{47.3}}(\textcolor{black}{+5.5}) & { \textbf{49.8}}(+6.4) & { \textbf{48.4}}(\textcolor{black}{+3.6}) & { \textbf{30.1}}(\textcolor{black}{+2.9}) \\ \bottomrule
\end{tabular}%
}
\vspace{-3mm}
\end{table*}

\subsection{Ablation Experiments}\label{subsection:ablation}

We conduct ablation experiments to analyze the effectiveness of each component in our proposed framework: Consistency Training (CT), Dynamic Cross-Entropy loss (DyCE), Vision-Language Pre-training (VLP), and Dense Language Guidance (DLG) in \autoref{tab:ablation}. 
\textbf{(A)} Starting with CT as our baseline, we observe moderate performance with mIoU scores of 54.5\% and 60.2\% on GTA5 and Synthia respectively with 100 labeled samples. \textbf{(B)} The addition of our DyCE loss significantly improves performance by addressing class imbalance issues, boosting the mIoU by 8.8\% (to 63.3\%) and 8.5\% (to 68.7\%) respectively. This, along with the findings in \autoref{tab:class-wise-comparison} and \autoref{tab:medical-comparison} demonstrates the effectiveness of the proposed dynamic loss function to address class imbalance.  
\textbf{(C)} When incorporating VLP without DyCE, we achieve better results than DyCE alone, with mIoU improvements of 11.1\% (to 65.6\%) and 11.7\% (to 71.9\%) over the baseline. This demonstrates the effectiveness of leveraging semantic knowledge from pre-trained vision-language models. \textbf{(D)} The addition of DLG further enhances performance substantially, reaching 70.3\% and 74.9\% mIoU, as it enables fine-grained semantic understanding through dense language embeddings.
\textbf{(E)} Finally, our full model combining all components achieves the best performance across all settings, with notable improvements of 16.6\% (to 71.1\%) and 16.7\% (to 76.9\%) over the baseline with 100 labeled samples. The consistent performance gains across different label ratios demonstrate the complementary nature of our proposed components in addressing the challenges of SSDA for semantic segmentation.

\begin{figure}
    \centering
    \includegraphics[width=\linewidth]{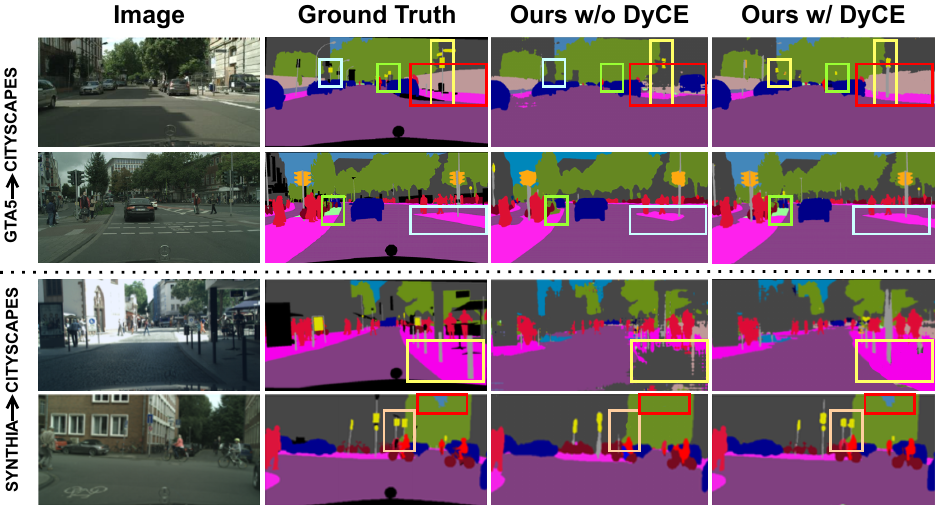}
    \caption{Qualitative segmentation performance of SemiDAViL with and without DyCE loss on 100 labeled target data.}
    \label{fig:qualitative-result}
    \vspace{-5mm}
\end{figure}

\subsection{Detailed Analysis on the DyCE Loss}\label{subsection:loss-validation}
We provide a comprehensive evaluation of our proposed method (with and without the DyCE loss) against existing UDA and SSDA methods that use various solutions on the class-imbalance problem on the GTA5 $\to$ Cityscapes and Synthia $\to$ Cityscapes benchmarks, using 100 labeled target samples in \autoref{tab:class-wise-comparison}. \textbf{(A)} Without DyCE, our model already achieves competitive mIoU scores, but the addition of DyCE enhances class separability by adaptively re-weighting underrepresented classes based on their occurrence, addressing inherent dataset imbalances. \textbf{(B)} The integration of DyCE loss shows significant improvements across a variety of challenging classes, particularly in tailed categories with high intra-class variation such as \textit{Fence}, \textit{Wall}, \textit{Terrain}, \textit{Rider}, and \textit{Pole}. We provide a qualitative visualization of improvement in segmentation performance by incorporating DyCE loss in \autoref{fig:qualitative-result}. 
For GTA5$\to$Cityscapes, the performance on tail classes like \textit{Wall} (51.9\%), \textit{Fence} (51.5\%), and \textit{Pole} (60.2\%) shows substantial improvement over previous state-of-the-art methods SS-ADA+U2PL \cite{yan2024ss} (47.1\%, 47.7\%, and 55.3\% respectively) and SS-ADA+UniMatch \cite{yan2024ss} (43.7\%, 45.1\%, and 53.3\% respectively). \textbf{(C)} We observe a significant improvement over UDA approaches like \cite{cardace2022shallow,li2022class,tranheden2021dacs}. \textbf{(D)} The overall mIoU rises from 70.3\% to 71.4\% by integrating DyCE loss, showing consistent performance gains across all label splits. Similar improvements are also evident in Synthia$\to$Cityscapes, demonstrating the robust superiority of our language-guided and class-balanced approach.

To further validate the effectiveness of DyCE loss, we evaluate it on a more challenging scenario: Synapse medical dataset with severe class imbalance (95.63\% background, 4.37\% foreground: foreground classes vary from 53.98\% to 0.14\%). As summarized in \autoref{tab:medical-comparison}, we highlight the significant impact of our DyCE loss as a plug-in enhancement across various SSL methods \cite{yu2019uncertainty,luo2022semi,wang2022debiased,wang2023dhc,wang2024towards}.
\textbf{(A)} For the lowest-performing general SSL method in \autoref{tab:medical-comparison}, i.e., UA-MT \cite{yu2019uncertainty}, its mDSC leaps from 20.3\% to 48.1\%, with minority classes like \textit{Gallbladder} improving from 0.0\% to 49.5\%, showcasing DyCE’s impressive adaptive weighting. Similarly, URPC's mDSC increases from 25.7\% to 48.9\%, whereas the previous best general SSL-based DePL sees a boost from 36.3\% to 51.8\%, underscoring DyCE’s capability to mitigate severe class imbalance by prioritizing underrepresented organs (e.g., \textit{Right Adrenal Gland} from 0.0\% to 26.9\%). \textbf{(B)} Recent SoTA of balanced SSL like DHC and A\&D also benefit, with DHC’s DSC rising from 48.6\% to 57.9\%, and A\&D reaching 65.5\% (up from 60.9\%). Notable gains include improvements in \textit{Gallbladder} (DHC: 66.0\% to 70.3\%) and \textit{Left Adrenal Gland} (A\&D: 27.2\% to 30.1\%), demonstrating DyCE’s effectiveness as a plug-in loss across class-imbalanced datasets. 
Further experimental findings, qualitative and quantitative analysis are provided in the \textbf{supplementary material}.

%% file: sec/5_conclusion.tex
\vspace{-2mm}
\section{Conclusion}\label{conclusion}
\vspace{-2mm}
In this work, we introduced SemiDAViL, a novel SSDA framework that leverages vision-language guidance and a dynamic loss formulation to address key challenges in domain-adaptive semantic segmentation. Through comprehensive experiments on several SSDA and SSL benchmarks, our method demonstrated consistent improvements over state-of-the-art techniques, especially in low-label regimes and class-imbalanced scenarios. The integration of vision-language pre-training, dense language embeddings, and the proposed DyCE loss contributes to discriminative feature extraction, better handling of minority classes, and enhanced semantic understanding. Overall, SemiDAViL sets a new benchmark in SSDA, showcasing strong generalizability across diverse domain shifts and label constraints.

%% file: sec/6_acknowledgement.tex
\section*{Acknowledgment}\label{acknowledge}

The authors greatly appreciate the financial support from the NSF project CMMI-2246673.

%% file: sec/X_suppl.tex
\begin{figure*}[t]
    \centering
    \begin{subfigure}[t]{\columnwidth}
        \centering
        \includegraphics[width=\columnwidth]{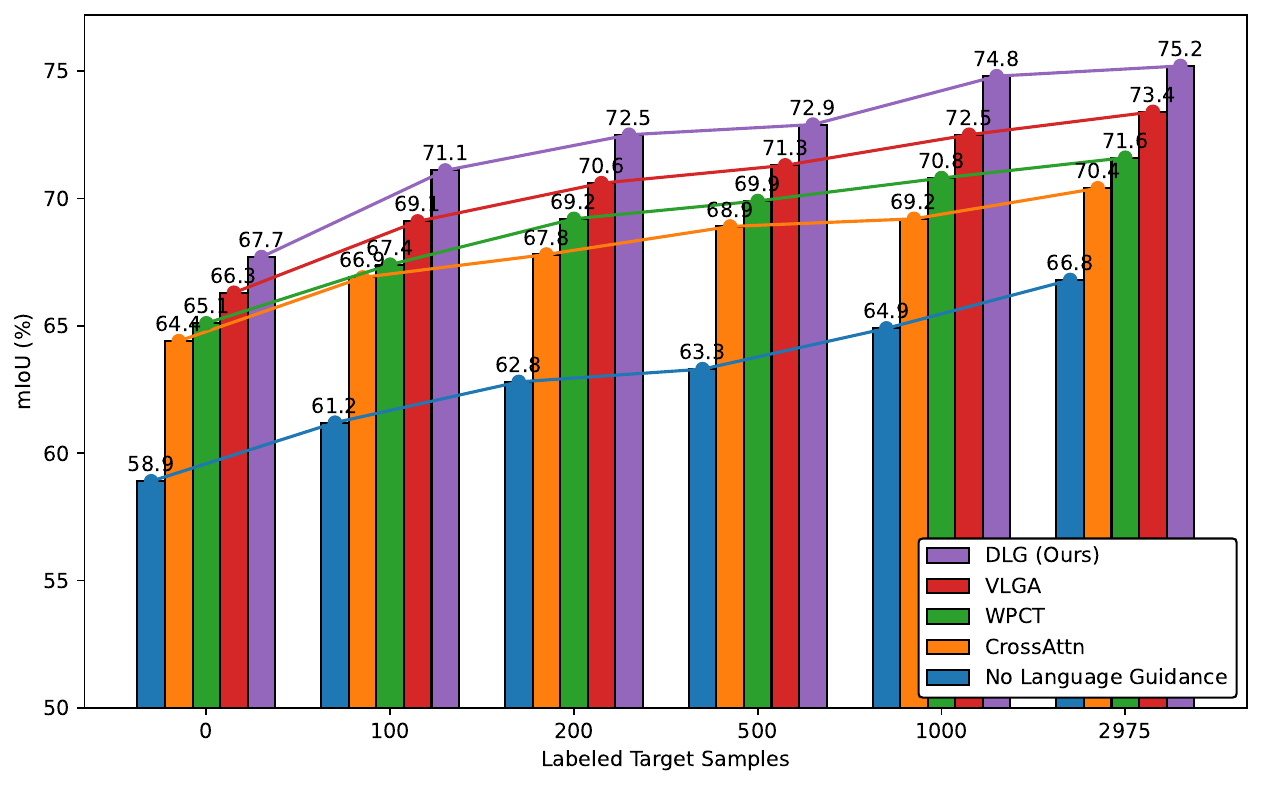}
        \caption{GTA5$\to$Cityscapes}
    \end{subfigure}%
    \begin{subfigure}[t]{\columnwidth}
        \centering
        \includegraphics[width=\columnwidth]{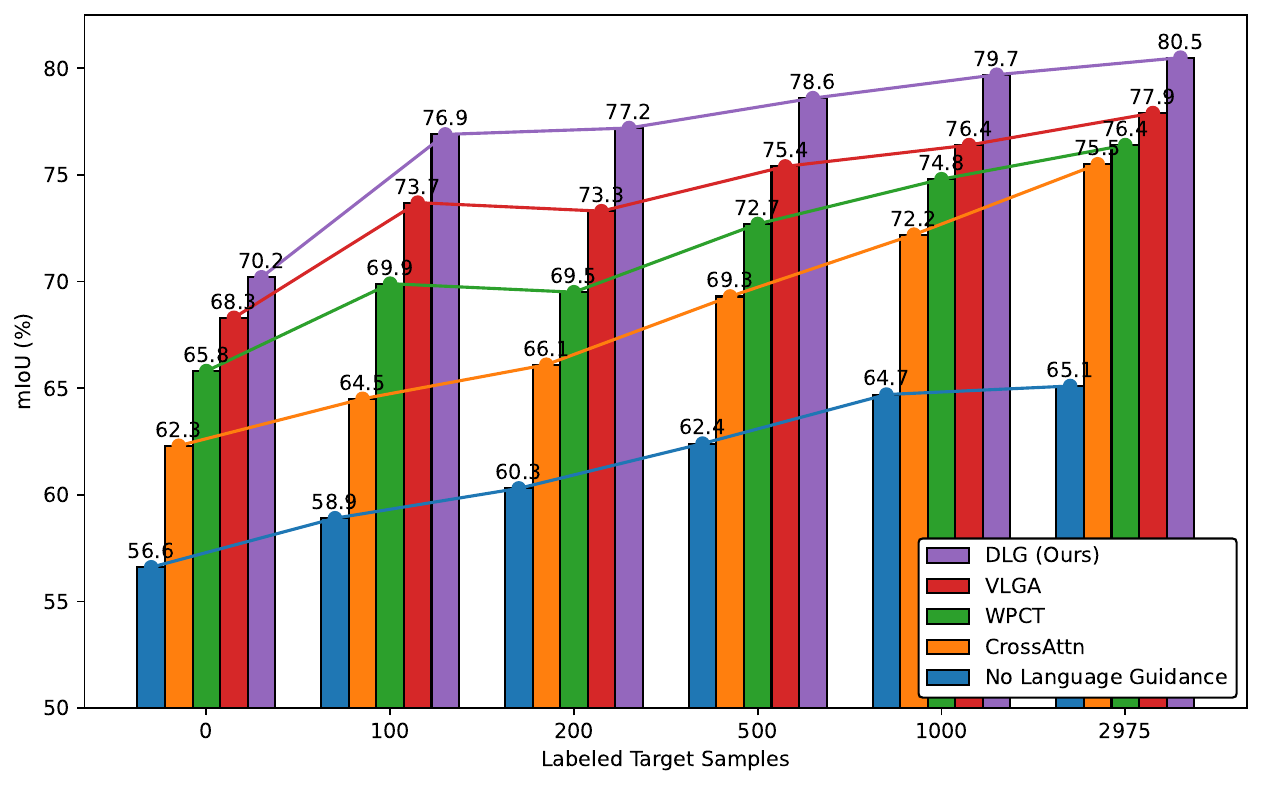}
        \caption{Synthia$\to$Cityscapes}
    \end{subfigure}
    \caption{Comparative analysis of multiple vision-language attention mechanisms: our Dense Language Guidance (DLG), VLGA \cite{hoyer2025semivl},  WPCT \cite{li2022language}, and Cross-attention \cite{chen2021crossvit} on (a) \textbf{GTA5$\to$Cityscapes} and (b) \textbf{Synthia$\to$Cityscapes} using different labeled target annotations. }
    \label{fig:attention-comparison-supple}
\end{figure*}

\begin{figure*}[t]
    \centering
    \begin{subfigure}[t]{\columnwidth}
        \centering
        \includegraphics[width=\columnwidth]{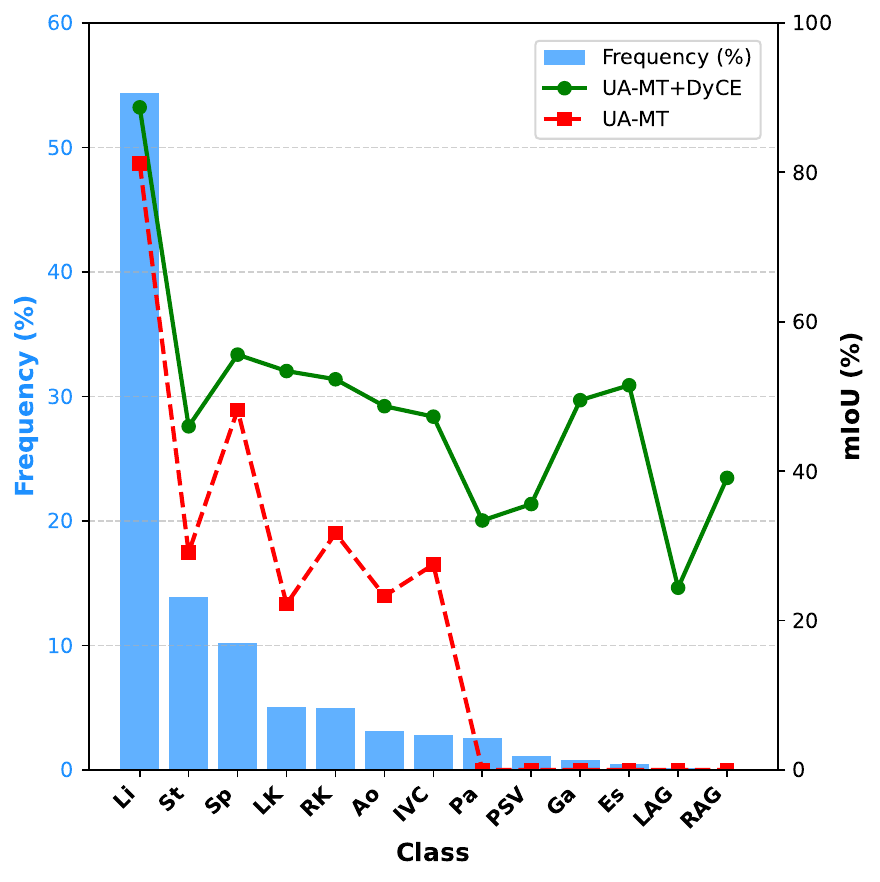}
        \caption{Improvement on UA-MT}
    \end{subfigure}%
    \begin{subfigure}[t]{\columnwidth}
        \centering
        \includegraphics[width=\columnwidth]{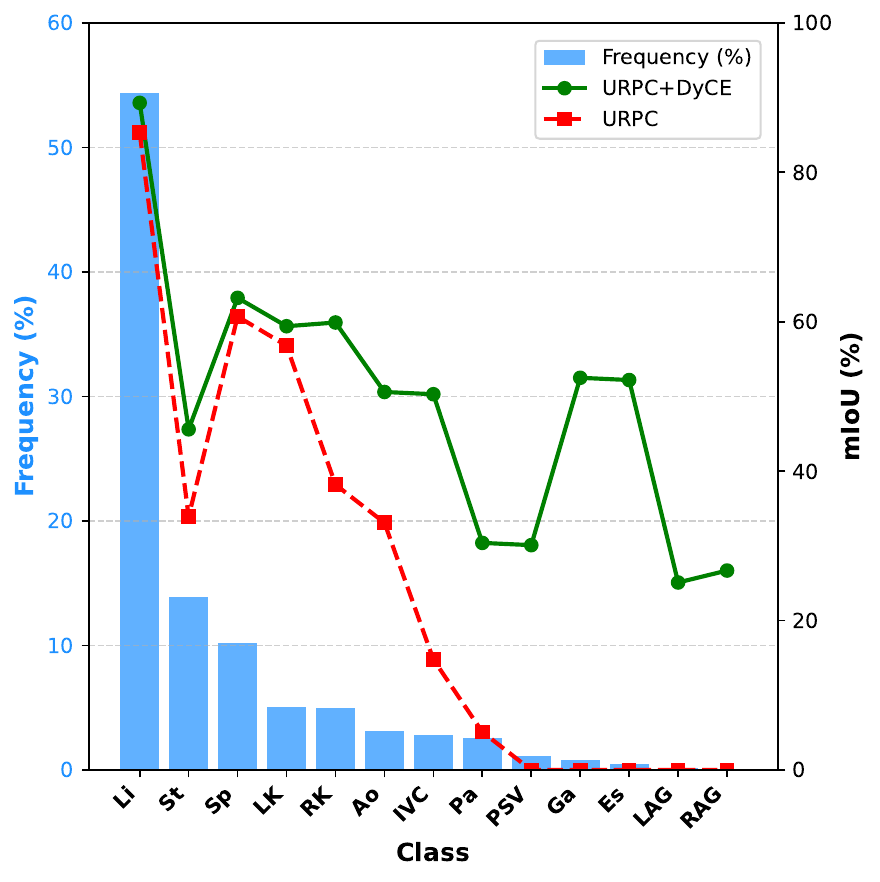}
        \caption{Improvement on URPC}
    \end{subfigure}
    \caption{Class distribution of the \textbf{Synapse} medical dataset and percentage improvement achieved by our DyCE loss in (a) UA-MT \cite{yu2019uncertainty} and (b) URPC \cite{luo2022semi} networks using 20\% labeled data in SSL setting.}
    \label{fig:synapse-stat-supple}
\end{figure*}

\begin{figure*}[t]
    \centering
    \begin{subfigure}[t]{\columnwidth}
        \centering
        \includegraphics[width=\columnwidth]{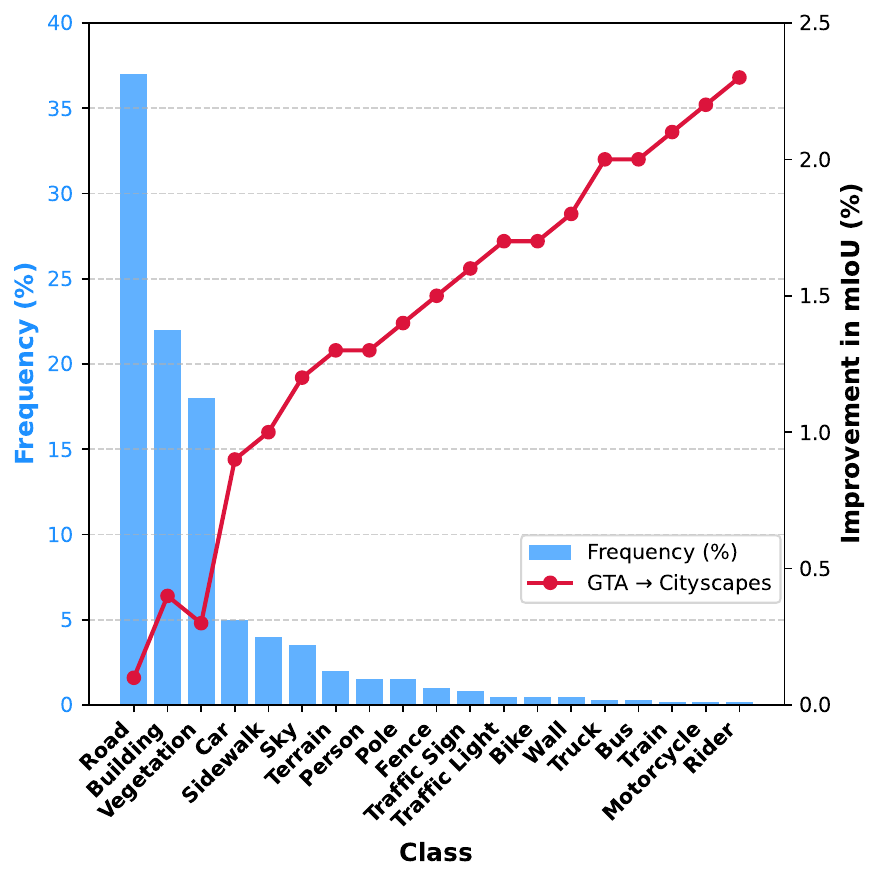}
        \caption{Improvement on \textbf{GTA5$\to$Cityscapes}}
    \end{subfigure}%
    \begin{subfigure}[t]{\columnwidth}
        \centering
        \includegraphics[width=\columnwidth]{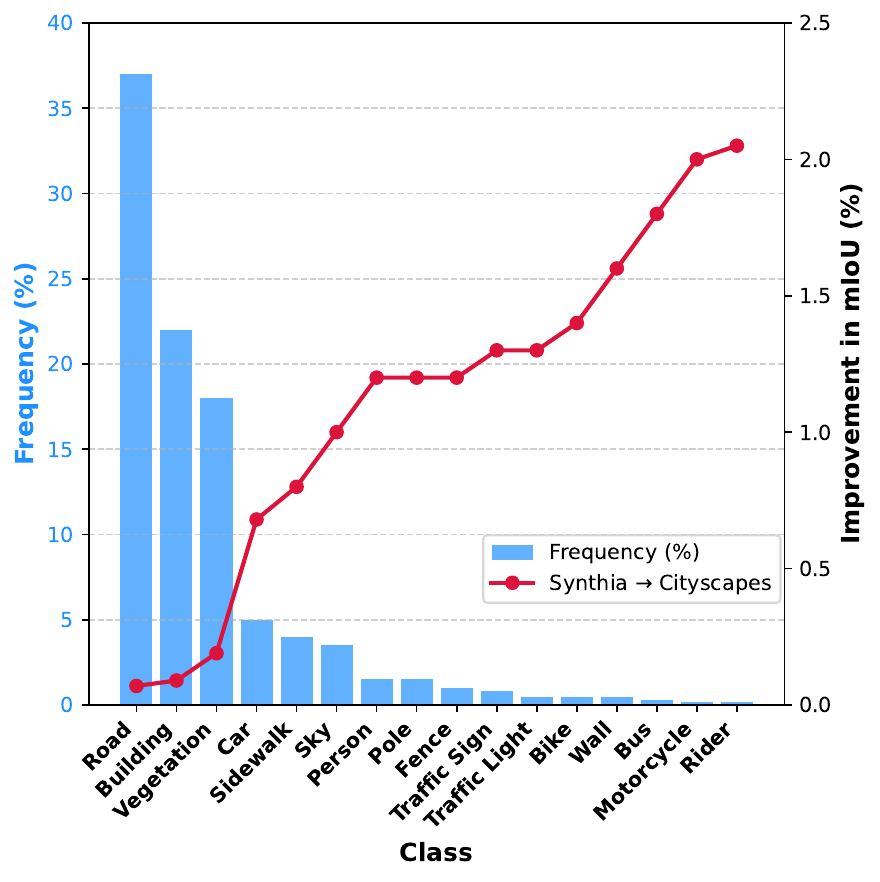}
        \caption{Improvement on \textbf{Synthia$\to$Cityscapes}}
    \end{subfigure}
    \caption{Class distribution of the Cityscapes dataset and percentage improvements achieved by our DyCE loss in (a) \textbf{GTA5$\to$Cityscapes} and (b) \textbf{Synthia$\to$Cityscapes} scenarios under the SSDA setting with 500 target annotations.}
    \label{fig:cityscapes-stat-supple}
\end{figure*}

\section{Overview}\label{overview-supple}
The supplementary material of SemiDAViL provides a detailed analysis of the superiority of our proposed vision-language attention mechanism in \autoref{effectiveness-dlg-supple}, studies the class statistics and analyzes the influence of DyCE loss for tail classes in \autoref{class-imbalance-supple}, detailed comparison with other class-balancing losses in \autoref{loss-comparison-supple}, qualitative analysis of SemiDAViL on segmentation tasks in \autoref{supple-qualitative}, and detailed comparison with SemiVL \cite{hoyer2025semivl} in \autoref{semivl-comparison}. Finally, we discuss some of the potential future directions in \autoref{sec:future-works}.



\section{Effectiveness of Dense Language Guidance}\label{effectiveness-dlg-supple}
To evaluate the effectiveness of our proposed vision-language attention mechanism using DLG, we perform a comparative analysis of DLG with some of the existing attention mechanisms. \autoref{fig:attention-comparison-supple} highlights the effectiveness of our proposed Dense Language Guidance (DLG) mechanism compared to previous vision-language integration approaches. DLG achieves state-of-the-art performance across all label settings in both GTA5 → Cityscapes and Synthia → Cityscapes benchmarks. For instance, in the zero-label setting, DLG attains a mean Intersection-over-Union (mIoU) of 67.7\% (GTA5) and 70.2\% (Synthia), outperforming the Word-Pixel Correlation Tensor (WPCT) \cite{li2022language} by +2.6\% and +4.4\%, respectively, and Vision-Language Guided Attention (VLGA) (which is used in SemiVL \cite{hoyer2025semivl}) by +1.4\% and +1.9\%. As the number of labeled target samples increases, DLG maintains consistent improvements, with mIoU gains of up to +3.6\% over WPCT and +2.6\% over VLGA at 2975 labeled samples. These results underscore DLG’s superior ability to fuse vision and language features effectively across various levels of supervision.

DLG's improved performance arises from its balanced treatment of vision and language features in its attention mechanism. Unlike prior methods that treat language features merely as attention weights, DLG transforms both visual and textual features into key-value pairs and treats them equivalently during cross-modal interaction. This enables a richer fusion process where attended vision and language features are integrated symmetrically, producing a true multimodal representation that retains information from both modalities. By normalizing and applying attention across both vision and language axes, DLG ensures comprehensive integration, avoiding the dominance of one modality over the other. The result is a robust mechanism capable of capturing nuanced cross-modal dependencies, which directly translates to improved generalization and segmentation performance, particularly in low-label scenarios where prior approaches struggle. Thus DLG sets a new benchmark for robust VL guidance in SSDA.

\section{Class Imbalance Analysis}\label{class-imbalance-supple}
We further evaluate the extent of class imbalance in Synapse and Cityscapes datasets in \autoref{fig:synapse-stat-supple} and \autoref{fig:cityscapes-stat-supple}, respectively along with the performance improvements brought by the proposed dynamic CE (DyCE) loss.
The plotted results in \autoref{fig:synapse-stat-supple} demonstrate a strong inverse correlation between class frequency and the efficacy of our DyCE loss in improving segmentation performance under SSL settings with 20\% annotations. For rare classes (e.g., Ga, Es, LAG, and RAG), which exhibit frequencies below 1\%, DyCE achieves substantial improvements, with URPC \cite{luo2022semi} and UA-MT \cite{yu2019uncertainty} backbones delivering gains exceeding 50\% in some cases. Conversely, for more frequently occurring classes like Li and St, DyCE exhibits modest gains, highlighting its ability to address class imbalance by prioritizing underrepresented categories. This trend underscores the robustness of DyCE in improving representation learning for minority classes, a critical challenge in medical image segmentation tasks.


The Cityscapes results in \autoref{fig:cityscapes-stat-supple} highlight the significant performance gains achieved for rare classes, such as Train, Motorcycle, and Rider, which collectively represent less than 1\% of the dataset. Notably, these classes exhibit improvements of up to 2.3\% (Rider) in the GTA5$\to$Cityscapes scenario and 2.05\% (Rider) in the Synthia$\to$Cityscapes scenario. Similarly, other infrequent classes, such as Bus, Truck, and Wall, also experience substantial boosts, ranging from 1.8\% to 2\% across both domain adaptation settings. This improvement underscores DyCE's effectiveness in addressing domain shift challenges for low-frequency classes, where traditional methods often struggle.
Conversely, the results for common classes like Road and Building, which dominate the dataset's frequency distribution, show relatively limited improvements of 0.1\% to 0.4\% for GTA5$\to$Cityscapes and 0.07\% to 0.09\% for Synthia$\to$Cityscapes. Classes with moderate frequency, such as Car, Sidewalk, and Vegetation, also show consistent but smaller gains, further confirming DyCE's capability to prioritize and balance underrepresented classes while maintaining stable performance for dominant categories. This class-specific focus allows the model to achieve improved overall segmentation quality without compromising accuracy in the more frequent classes.

Overall, DyCE demonstrates superior efficacy in addressing class imbalance by dynamically prioritizing underrepresented tail classes, resulting in significant performance gains for rare categories while maintaining stability for dominant ones, establishing its critical role in advancing segmentation tasks where tail-class accuracy is crucial.


\begin{figure*}
    \centering
    \includegraphics[width=\linewidth]{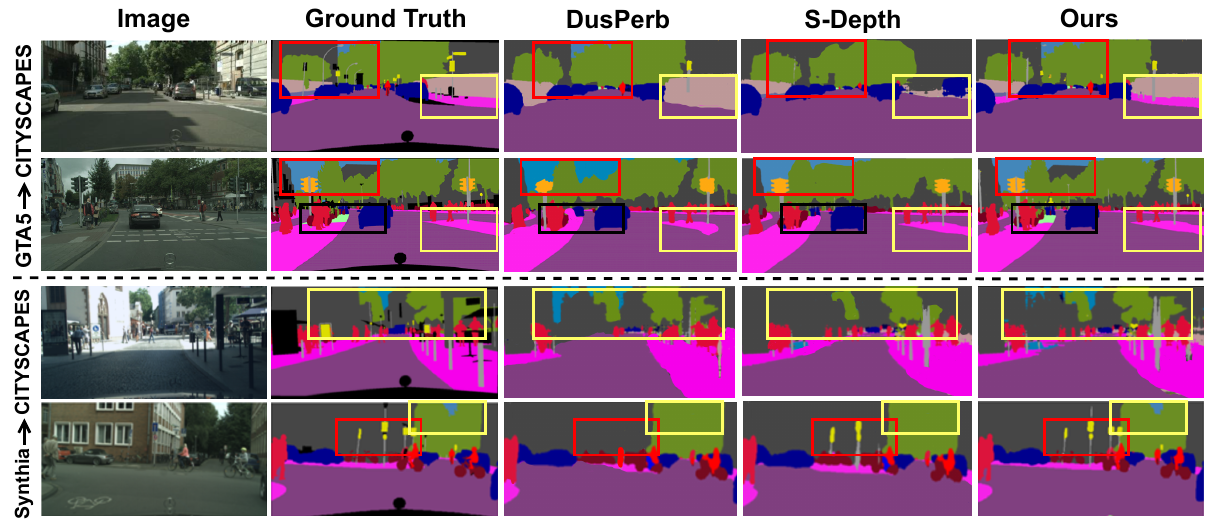}
    \caption{Qualitative comparison of SemiDAViL with previous state-of-the-art method, DusPerb \cite{yang2023revisiting}, S-Depth \cite{hoyer2023improving} on 100-labeled target data on \textbf{GTA5$\to$Cityscapes} and \textbf{Synthia$\to$Cityscapes} adaptation settings. }
    \label{fig:supple-qualitative-comparison}
\end{figure*}

\section{Comparison with Class-balancing Losses}\label{loss-comparison-supple}

\begin{table}[tbp]
\centering
\caption{Performance comparison of our DyCE loss with previous loss functions to address class imbalance. We report 19-class and 16-class mIoU scores for the \textbf{GTA5$\to$Cityscapes} and \textbf{Synthia$\to$Cityscapes} settings, respectively across 0, 100, 200, 500, 100, and 2975 (100\%) labeled target images. Our results are \textbf{highlighted} whereas the previous-best and second-best results are marked in \textcolor{red}{red} and \textcolor{blue}{blue}. }
\label{tab:loss-supple}
\resizebox{0.8\columnwidth}{!}{%
\begin{tabular}{@{}ccccccc@{}}
\toprule
\multicolumn{7}{c}{\textbf{GTA5$\to$Cityscapes}} \\ \midrule
 & \multicolumn{6}{c}{\textbf{Labeled Target Samples}} \\ \cmidrule(l){2-7} 
\multirow{-2}{*}{\textbf{Loss}} & \textbf{0} & \textbf{100} & \textbf{200} & \textbf{500} & \textbf{1000} & \textbf{2975} \\ \midrule
CE \cite{zhang2018generalized}& 66.9 & 70.3 & 71.6 & 72.1 & 73.9 & 74.4 \\
WCE \cite{aurelio2019learning}& 64.3 & 69.4 & 70.5 & 71.8 & 72.7 & 73.3 \\
DCE \cite{li2019dual}& {\color[HTML]{3531FF} 67.1} & 70.5 & 71.6 & {\color[HTML]{FE0000} 72.3} & 73.3 & 74.7 \\
FL \cite{lin2017focal}& 67.0 & {\color[HTML]{3531FF} 70.4} & {\color[HTML]{3531FF} 71.8} & {\color[HTML]{3531FF} 72.2} & {\color[HTML]{FE0000} 74.1} & {\color[HTML]{FE0000} 74.9} \\
DFL \cite{hossain2021dual}& {\color[HTML]{FE0000} 67.3} & {\color[HTML]{FE0000} 70.6} & {\color[HTML]{FE0000} 71.9} & {\color[HTML]{3531FF} 72.2} & {\color[HTML]{3531FF} 74.0} & {\color[HTML]{3531FF} 74.8} \\ \hline
\textbf{DyCE (ours)} & \textbf{67.7} & \textbf{71.1} & \textbf{72.5} & \textbf{72.9} & \textbf{74.8} & \textbf{75.2} \\ \midrule
\multicolumn{7}{c}{\textbf{Synthia$\to$Cityscapes}} \\ \midrule
CE \cite{zhang2018generalized}& 69.5 & 74.9 & 76.8 & 77.7 & {\color[HTML]{3531FF} 79.2} & 79.6 \\
WCE \cite{aurelio2019learning}& 67.2 & 73.1 & 74.4 & 75.3 & 77.6 & 78.9 \\
DCE \cite{li2019dual}& 69.7 & 75.2 & 76.8 & {\color[HTML]{3531FF} 78.1} & 79.1 & {\color[HTML]{3531FF} 79.7} \\
FL \cite{lin2017focal}& {\color[HTML]{3531FF} 69.8} & {\color[HTML]{3531FF} 75.4} & {\color[HTML]{3531FF} 76.9} & 78.0 & {\color[HTML]{FE0000} 79.3} & {\color[HTML]{3531FF} 79.7} \\
DFL\cite{hossain2021dual} & {\color[HTML]{FE0000} 69.9} & {\color[HTML]{FE0000} 75.7} & {\color[HTML]{FE0000} 77.0} & {\color[HTML]{FE0000} 78.2} & 78.9 & {\color[HTML]{FE0000} 79.8} \\ \hline
\textbf{DyCE (ours)} & \textbf{70.2} & \textbf{76.9} & \textbf{77.2} & \textbf{78.6} & \textbf{79.7} & \textbf{80.5} \\ \bottomrule
\end{tabular}%
}
\end{table}

In this section, we present a comparative analysis of our proposed Dynamic Cross Entropy (DyCE) loss against several state-of-the-art loss functions designed to address class imbalance in semantic segmentation tasks. The methods under comparison include Cross Entropy (CE) \cite{zhang2018generalized} , Weighted Cross Entropy (WCE) \cite{aurelio2019learning}, Focal Loss (FL) \cite{lin2017focal}, Dual Cross Entropy (DCE) \cite{li2019dual}, and Dual Focal Loss (DFL)  \citet{hossain2021dual}. 

Cross Entropy serves as a baseline, calculating the logarithmic difference between predicted and actual class distributions. Weighted Cross Entropy introduces class-specific weights to emphasize underrepresented categories. Focal Loss modulates the loss contribution of well-classified examples, allowing the model to focus on challenging instances. Dual Cross Entropy incorporates a regularization term to balance positive and negative predictions. Dual Focal Loss combines adaptive scaling with regularization to achieve more balanced gradient propagation. Our proposed DyCE loss builds upon these foundations, introducing a dynamic, gradient-adaptive mechanism to optimize convergence and segmentation performance across varying levels of supervision.

\autoref{tab:loss-supple} establishes the preeminence of DyCE in mitigating class imbalance across varying degrees of target annotations. In the \textbf{GTA$\to$Cityscapes} domain adaptation scenario, DyCE achieves a 19-class mIoU of 67.7\% in the absence of labeled target samples, scaling up to 75.2\% under full supervision (2975 labels). For \textbf{Synthia$\to$Cityscapes}, DyCE exhibits analogous dominance, attaining a 16-class mIoU of 70.2\% without annotations and 80.5\% under complete supervision. These findings underscore DyCE's adaptability and robustness across supervision levels. Unlike WCE, which employs static weighting, and DCE, which incorporates a regularization term, DyCE dynamically accentuates harder-to-classify categories without succumbing to gradient attenuation. This design enables it to surpass state-of-the-art methods such as DFL and FL, particularly under annotation-scarce regimes.

CE establishes a baseline yet falters in addressing class imbalance due to uniform loss weighting. WCE refines this by imposing fixed penalties on minority classes; however, static weights remain suboptimal in evolving scenarios. FL introduces dynamic scaling tailored to class difficulty, excelling in emphasizing hard-to-classify examples, but its susceptibility to vanishing gradients impedes training convergence. DCE mitigates this gradient vanishing by regularizing negative class predictions, albeit at the cost of disproportionately penalizing false negatives. DFL synthesizes FL’s adaptive scaling with DCE’s regularization, ensuring balanced gradient propagation. DyCE advances this paradigm through a gradient-adaptive mechanism that dynamically modulates loss and penalizes misclassifications more rigorously, especially for underrepresented classes. This innovative construct ensures optimal convergence and superior segmentation performance, as evidenced by DyCE's consistent outperformance across all experimental scenarios in \autoref{tab:loss-supple}.

\section{Qualitative Results}\label{supple-qualitative}

We provide a qualitative comparison of our proposed SemiDAViL with the previous best methods, DusPerb \cite{yang2023revisiting}, S-Depth \cite{hoyer2023improving}, and the available ground truth label in \autoref{fig:supple-qualitative-comparison}. Experiments are performed using 100-labeled target samples on \textbf{GTA5$\to$Cityscapes} and \textbf{Synthia$\to$Cityscapes} settings. 
Upon closer examination, the proposed method consistently produces segmentation masks that more closely align with the ground truth than the other two methods. In particular, our method demonstrates superior performance in accurately delineating object boundaries and capturing fine details for semantically confusing classes (e.g., sidewalk vs. wall, rider vs bike, traffic light vs. vegetation, wall vs. sky, etc.), owing to the strong semantic prior from vision-language initialization and dense language guidance using multimodal attention. This suggests that the proposed method has a better capacity for understanding and interpreting the relationship between linguistic descriptions and visual features, resulting in more accurate and refined segmentation outputs. Moreover, our approach shows a marked improvement for tail classes (e.g., rider, motorcycle, wall, etc.) in precisely segmenting intricate shapes and maintaining object integrity. This improvement can be attributed to our proposed class-balancing DyCE loss, which dynamically prioritizes imbalanced and underperforming classes. Previous SoTA methods like \cite{yang2023revisiting, hoyer2023improving} fall short in these two major aspects, leading to suboptimal performance.

\section{Comparison with SemiVL}\label{semivl-comparison}
Although they may appear similar at first glance, our work is fundamentally different from SemiVL \cite{hoyer2025semivl}. Whereas SemiVL targets semi-supervised semantic segmentation with a focus on label efficiency—employing a language-guided decoder that leverages frozen CLIP predictions and dataset-specific class definitions—our method, SemiDAViL, pioneers semi-supervised domain adaptation. We address the domain shift challenge by integrating a Dense Language Guidance (DLG) module that fuses fine-grained visual and textual embeddings for robust, pixel-level semantic alignment across domains. Furthermore, our approach tackles class imbalance through a novel Dynamic Cross-Entropy (DyCE) loss that reweights minority classes during training. In addition, our pseudo-labeling strategy synergistically combines consistency regularization with dense language embeddings to refine predictions, while our language guidance utilizes detailed captions for both content and spatial positioning rather than fixed class definitions. Together, these innovations enable SemiDAViL to effectively bridge the semantic gap between source and target domains, setting it apart as the first language-guided semi-supervised DA method for semantic segmentation.

To further validate the differences quantitatively, we perform experiments under both semi-supervised domain adaptation (SSDA) and semi-supervised learning (SSL) settings using varying numbers of labeled target samples. Under the SSDA scenario—specifically adapting from GTA or synthetic data to Cityscapes—our method consistently achieves higher mIoU scores across all target sample sizes. For example, when adapting from Syn. to Cityscapes, our method attains 76.9, 77.2, 78.6, and 79.7 mIoU with 100, 200, 500, and 1000 labeled samples respectively, outperforming SemiVL by up to 5.5 mIoU. This performance gain is primarily attributed to our integration of domain adaptation components, such as the Dense Language Guidance (DLG) module, which fuses visual and textual features to better align semantic representations across domains, and the novel Dynamic Cross-Entropy (DyCE) loss that rebalances class distributions to mitigate source bias. In contrast, SemiVL, which lacks explicit domain adaptation mechanisms, tends to overfit to the abundant source labels, resulting in a suboptimal adaptation to the target domain. Moreover, even under the SSL setting (Cityscapes→Cityscapes), where domain shift is not a factor, our method still outperforms SemiVL (81.6 vs. 80.4 mIoU at 1000 labeled samples), indicating that our approach enhances feature localization and pseudo-label refinement through a more robust consistency training framework. These detailed experimental results confirm that our method addresses domain shift more effectively and improves the overall semantic segmentation performance under limited annotation.

\begin{table}[htbp]
    \centering
    \caption{Quantitative comparison with SemiVL \cite{hoyer2025semivl} on SSDA and SSL settings.}
    \resizebox{0.8\columnwidth}{!}{%
    \begin{tabular}{@{}cccccc@{}}
    \toprule
    \multirow{2}{*}{\textbf{Type}} & \multirow{2}{*}{\textbf{Method}} & \multicolumn{4}{c}{\textbf{Labeled Target Sample}} \\ \cmidrule(l){3-6} 
     &  & \textbf{100} & \textbf{200} & \textbf{500} & \textbf{1000} \\ \midrule
    \multirow{2}{*}{\begin{tabular}[c]{@{}c@{}}SSDA \\ (GTA$\to$City.)\end{tabular}} & SemiVL \cite{hoyer2025semivl} & 68.5 & 69.9 & 70.6 & 71.7 \\
     & Ours & 71.1 & 72.5 & 72.9 & 74.8 \\ \midrule
    \multirow{2}{*}{\begin{tabular}[c]{@{}c@{}}SSDA \\ (Syn.$\to$City.)\end{tabular}} & SemiVL \cite{hoyer2025semivl} & 71.4 & 72.3 & 74.3 & 75.1 \\
     & Ours & \textbf{76.9} & \textbf{77.2} & \textbf{78.6} & \textbf{79.7} \\ \bottomrule
     \multirow{2}{*}{\begin{tabular}[c]{@{}c@{}}SSL \\ (City.$\to$City.)\end{tabular}} & SemiVL \cite{hoyer2025semivl} & 76.2 & 77.9 & 80.2 & 80.4 \\
     & Ours & \textbf{77.1 }& \textbf{78.2} & \textbf{81.4} & \textbf{81.6} \\ \bottomrule
    \end{tabular}%
    }    
    \label{tab:my_label}
\end{table}

\section{Future Works}\label{sec:future-works}
While SemiDAViL demonstrates strong performance in mitigating class imbalance and enhancing segmentation accuracy, there remain areas that offer opportunities for further refinement. The reliance on pre-trained vision-language models like CLIP may pose challenges in adapting to domains where such resources are limited or less aligned with the data. The dense multimodal attention and dynamic loss modulation, while effective, could benefit from optimizations to ensure scalability across larger datasets and real-time applications. The use of off-the-shelf captioning models in Dense Language Guidance (DLG) highlights the importance of high-quality linguistic features, which could be further enhanced for greater robustness. Addressing these aspects could unlock even broader applications and performance improvements for this promising approach.

%% file: paper.bbl
\begin{thebibliography}{109}
\providecommand{\natexlab}[1]{#1}
\providecommand{\url}[1]{\texttt{#1}}
\expandafter\ifx\csname urlstyle\endcsname\relax
  \providecommand{\doi}[1]{doi: #1}\else
  \providecommand{\doi}{doi: \begingroup \urlstyle{rm}\Url}\fi

\bibitem[Aurelio et~al.(2019)Aurelio, De~Almeida, de~Castro, and Braga]{aurelio2019learning}
Yuri~Sousa Aurelio, Gustavo~Matheus De~Almeida, Cristiano~Leite de Castro, and Antonio~Padua Braga.
\newblock Learning from imbalanced data sets with weighted cross-entropy function.
\newblock \emph{Neural processing letters}, 50:\penalty0 1937--1949, 2019.

\bibitem[Basak and Yin(2024)]{basak2024quest}
Hritam Basak and Zhaozheng Yin.
\newblock Quest for clone: Test-time domain adaptation for medical image segmentation by searching the closest clone in latent space.
\newblock In \emph{International Conference on Medical Image Computing and Computer-Assisted Intervention}, pages 555--566. Springer, 2024.

\bibitem[Basak and Yin(2025)]{basak2025forget}
Hritam Basak and Zhaozheng Yin.
\newblock Forget more to learn more: Domain-specific feature unlearning for semi-supervised and unsupervised domain adaptation.
\newblock In \emph{European Conference on Computer Vision}, pages 130--148. Springer, 2025.

\bibitem[Basak et~al.(2022)Basak, Ghosal, and Sarkar]{basak2022addressing}
Hritam Basak, Sagnik Ghosal, and Ram Sarkar.
\newblock Addressing class imbalance in semi-supervised image segmentation: A study on cardiac mri.
\newblock In \emph{International Conference on Medical Image Computing and Computer-Assisted Intervention}, pages 224--233. Springer, 2022.

\bibitem[Cardace et~al.(2022)Cardace, Ramirez, Salti, and Di~Stefano]{cardace2022shallow}
Adriano Cardace, Pierluigi~Zama Ramirez, Samuele Salti, and Luigi Di~Stefano.
\newblock Shallow features guide unsupervised domain adaptation for semantic segmentation at class boundaries.
\newblock In \emph{Proceedings of the IEEE/CVF Winter Conference on Applications of Computer Vision}, pages 1160--1170, 2022.

\bibitem[Chen et~al.(2022{\natexlab{a}})Chen, Jiang, Wang, Wan, Wang, and Long]{chen2022debiased}
Baixu Chen, Junguang Jiang, Ximei Wang, Pengfei Wan, Jianmin Wang, and Mingsheng Long.
\newblock Debiased self-training for semi-supervised learning.
\newblock \emph{Advances in Neural Information Processing Systems}, 35:\penalty0 32424--32437, 2022{\natexlab{a}}.

\bibitem[Chen et~al.(2021{\natexlab{a}})Chen, Fan, and Panda]{chen2021crossvit}
Chun-Fu~Richard Chen, Quanfu Fan, and Rameswar Panda.
\newblock Crossvit: Cross-attention multi-scale vision transformer for image classification.
\newblock In \emph{Proceedings of the IEEE/CVF international conference on computer vision}, pages 357--366, 2021{\natexlab{a}}.

\bibitem[Chen et~al.(2022{\natexlab{b}})Chen, Fan, Wang, Wang, Schiele, Xie, Savvides, and Raj]{chen2022embarrassingly}
Hao Chen, Yue Fan, Yidong Wang, Jindong Wang, Bernt Schiele, Xing Xie, Marios Savvides, and Bhiksha Raj.
\newblock An embarrassingly simple baseline for imbalanced semi-supervised learning.
\newblock \emph{arXiv preprint arXiv:2211.11086}, 2022{\natexlab{b}}.

\bibitem[Chen et~al.(2021{\natexlab{b}})Chen, Li, Bai, Yang, Jiang, and Miao]{chen2021review}
Leiyu Chen, Shaobo Li, Qiang Bai, Jing Yang, Sanlong Jiang, and Yanming Miao.
\newblock Review of image classification algorithms based on convolutional neural networks.
\newblock \emph{Remote Sensing}, 13\penalty0 (22):\penalty0 4712, 2021{\natexlab{b}}.

\bibitem[Chen et~al.(2017)Chen, Papandreou, Kokkinos, Murphy, and Yuille]{chen2017deeplab}
Liang-Chieh Chen, George Papandreou, Iasonas Kokkinos, Kevin Murphy, and Alan~L Yuille.
\newblock Deeplab: Semantic image segmentation with deep convolutional nets, atrous convolution, and fully connected crfs.
\newblock \emph{IEEE transactions on pattern analysis and machine intelligence}, 40\penalty0 (4):\penalty0 834--848, 2017.

\bibitem[Chen et~al.(2021{\natexlab{c}})Chen, Jia, He, Shi, and Liu]{chen2021semidual}
Shuaijun Chen, Xu Jia, Jianzhong He, Yongjie Shi, and Jianzhuang Liu.
\newblock Semi-supervised domain adaptation based on dual-level domain mixing for semantic segmentation.
\newblock In \emph{Proceedings of the IEEE/CVF Conference on Computer Vision and Pattern Recognition}, pages 11018--11027, 2021{\natexlab{c}}.

\bibitem[Chen et~al.(2021{\natexlab{d}})Chen, Yuan, Zeng, and Wang]{chen2021semi}
Xiaokang Chen, Yuhui Yuan, Gang Zeng, and Jingdong Wang.
\newblock Semi-supervised semantic segmentation with cross pseudo supervision.
\newblock In \emph{Proceedings of the IEEE/CVF conference on computer vision and pattern recognition}, pages 2613--2622, 2021{\natexlab{d}}.

\bibitem[Chen et~al.(2021{\natexlab{e}})Chen, Yuan, Zeng, and Wang]{chen2021semicps}
Xiaokang Chen, Yuhui Yuan, Gang Zeng, and Jingdong Wang.
\newblock Semi-supervised semantic segmentation with cross pseudo supervision.
\newblock In \emph{Proceedings of the IEEE/CVF conference on computer vision and pattern recognition}, pages 2613--2622, 2021{\natexlab{e}}.

\bibitem[Chen et~al.(2021{\natexlab{f}})Chen, Ouyang, Zhu, and Agam]{chen2021complexmix}
Ying Chen, Xu Ouyang, Kaiyue Zhu, and Gady Agam.
\newblock Semi-supervised domain adaptation for semantic segmentation.
\newblock \emph{arXiv preprint arXiv:2110.10639}, 2021{\natexlab{f}}.

\bibitem[Chen et~al.(2022{\natexlab{c}})Chen, Ouyang, Zhu, and Agam]{chen2022semi}
Ying Chen, Xu Ouyang, Kaiyue Zhu, and Gady Agam.
\newblock Semi-supervised dual-domain adaptation for semantic segmentation.
\newblock In \emph{2022 26th International Conference on Pattern Recognition (ICPR)}, pages 230--237. IEEE, 2022{\natexlab{c}}.

\bibitem[Cordts et~al.(2016)Cordts, Omran, Ramos, Rehfeld, Enzweiler, Benenson, Franke, Roth, and Schiele]{cordts2016cityscapes}
Marius Cordts, Mohamed Omran, Sebastian Ramos, Timo Rehfeld, Markus Enzweiler, Rodrigo Benenson, Uwe Franke, Stefan Roth, and Bernt Schiele.
\newblock The cityscapes dataset for semantic urban scene understanding.
\newblock In \emph{Proceedings of the IEEE conference on computer vision and pattern recognition}, pages 3213--3223, 2016.

\bibitem[Deng et~al.(2009)Deng, Dong, Socher, Li, Li, and Fei-Fei]{deng2009imagenet}
Jia Deng, Wei Dong, Richard Socher, Li-Jia Li, Kai Li, and Li Fei-Fei.
\newblock Imagenet: A large-scale hierarchical image database.
\newblock In \emph{2009 IEEE conference on computer vision and pattern recognition}, pages 248--255. Ieee, 2009.

\bibitem[Ding et~al.(2021)Ding, Liu, Wang, and Jiang]{ding2021vision}
Henghui Ding, Chang Liu, Suchen Wang, and Xudong Jiang.
\newblock Vision-language transformer and query generation for referring segmentation.
\newblock In \emph{Proceedings of the IEEE/CVF International Conference on Computer Vision}, pages 16321--16330, 2021.

\bibitem[Ding et~al.(2022)Ding, Xue, Xia, and Dai]{ding2022decoupling}
Jian Ding, Nan Xue, Gui-Song Xia, and Dengxin Dai.
\newblock Decoupling zero-shot semantic segmentation.
\newblock In \emph{Proceedings of the IEEE/CVF Conference on Computer Vision and Pattern Recognition}, pages 11583--11592, 2022.

\bibitem[Dosovitskiy et~al.(2021)Dosovitskiy, Beyer, Kolesnikov, Weissenborn, Zhai, Unterthiner, Dehghani, Minderer, Heigold, Gelly, Uszkoreit, and Houlsby]{dosovitskiy2020vit}
Alexey Dosovitskiy, Lucas Beyer, Alexander Kolesnikov, Dirk Weissenborn, Xiaohua Zhai, Thomas Unterthiner, Mostafa Dehghani, Matthias Minderer, Georg Heigold, Sylvain Gelly, Jakob Uszkoreit, and Neil Houlsby.
\newblock An image is worth 16x16 words: Transformers for image recognition at scale.
\newblock \emph{ICLR}, 2021.

\bibitem[Erkent and Laugier(2020)]{erkent2020semantic}
{\"O}zg{\"u}r Erkent and Christian Laugier.
\newblock Semantic segmentation with unsupervised domain adaptation under varying weather conditions for autonomous vehicles.
\newblock \emph{IEEE Robotics and Automation Letters}, 5\penalty0 (2):\penalty0 3580--3587, 2020.

\bibitem[Fahes et~al.(2024)Fahes, Vu, Bursuc, P{\'e}rez, and de~Charette]{fahes2024simple}
Mohammad Fahes, Tuan-Hung Vu, Andrei Bursuc, Patrick P{\'e}rez, and Raoul de Charette.
\newblock A simple recipe for language-guided domain generalized segmentation.
\newblock In \emph{Proceedings of the IEEE/CVF Conference on Computer Vision and Pattern Recognition}, pages 23428--23437, 2024.

\bibitem[Feng et~al.(2022)Feng, Zhou, Gu, Tan, Cheng, Lu, Shi, and Ma]{feng2022dmt}
Zhengyang Feng, Qianyu Zhou, Qiqi Gu, Xin Tan, Guangliang Cheng, Xuequan Lu, Jianping Shi, and Lizhuang Ma.
\newblock Dmt: Dynamic mutual training for semi-supervised learning.
\newblock \emph{Pattern Recognition}, 130:\penalty0 108777, 2022.

\bibitem[French et~al.(2020)French, Aila, Laine, Mackiewicz, and Finlayson]{french2019semi}
Geoff French, Timo Aila, Samuli Laine, Michal Mackiewicz, and Graham Finlayson.
\newblock Semi-supervised semantic segmentation needs strong, high-dimensional perturbations.
\newblock \emph{BMVC}, 2020.

\bibitem[Fu et~al.(2024)Fu, Nie, Li, Lin, Wu, Li, Wang, Liu, and Wang]{fu2024iidminterintradomainmixing}
Weifu Fu, Qiang Nie, Jialin Li, Yuhuan Lin, Kai Wu, Jian Li, Yabiao Wang, Yong Liu, and Chengjie Wang.
\newblock Iidm: Inter and intra-domain mixing for semi-supervised domain adaptation in semantic segmentation, 2024.

\bibitem[Gao et~al.(2024)Gao, Wang, and Zhang]{gao2024delve}
Yuan Gao, Zilei Wang, and Yixin Zhang.
\newblock Delve into source and target collaboration in semi-supervised domain adaptation for semantic segmentation.
\newblock In \emph{2024 IEEE International Conference on Multimedia and Expo (ICME)}, pages 1--6. IEEE, 2024.

\bibitem[Ghiasi et~al.(2022)Ghiasi, Gu, Cui, and Lin]{ghiasi2022scaling}
Golnaz Ghiasi, Xiuye Gu, Yin Cui, and Tsung-Yi Lin.
\newblock Scaling open-vocabulary image segmentation with image-level labels.
\newblock In \emph{European Conference on Computer Vision}, pages 540--557. Springer, 2022.

\bibitem[Griffin et~al.(2022)Griffin, Holub, and Perona]{griffin_holub_perona_2022}
Gregory Griffin, Alex Holub, and Pietro Perona.
\newblock Caltech 256, 2022.

\bibitem[Guo and Li(2022)]{guo2022class}
Lan-Zhe Guo and Yu-Feng Li.
\newblock Class-imbalanced semi-supervised learning with adaptive thresholding.
\newblock In \emph{International conference on machine learning}, pages 8082--8094. PMLR, 2022.

\bibitem[He et~al.(2020)He, Fan, Wu, Xie, and Girshick]{he2020momentum}
Kaiming He, Haoqi Fan, Yuxin Wu, Saining Xie, and Ross Girshick.
\newblock Momentum contrast for unsupervised visual representation learning.
\newblock In \emph{Proceedings of the IEEE/CVF conference on computer vision and pattern recognition}, pages 9729--9738, 2020.

\bibitem[Hossain et~al.(2021)Hossain, Betts, and Paplinski]{hossain2021dual}
Md~Sazzad Hossain, John~M Betts, and Andrew~P Paplinski.
\newblock Dual focal loss to address class imbalance in semantic segmentation.
\newblock \emph{Neurocomputing}, 462:\penalty0 69--87, 2021.

\bibitem[Hoyer et~al.(2022{\natexlab{a}})Hoyer, Dai, and Van~Gool]{hoyer2022daformer}
Lukas Hoyer, Dengxin Dai, and Luc Van~Gool.
\newblock Daformer: Improving network architectures and training strategies for domain-adaptive semantic segmentation.
\newblock In \emph{Proceedings of the IEEE/CVF conference on computer vision and pattern recognition}, pages 9924--9935, 2022{\natexlab{a}}.

\bibitem[Hoyer et~al.(2022{\natexlab{b}})Hoyer, Dai, and Van~Gool]{hoyer2022hrda}
Lukas Hoyer, Dengxin Dai, and Luc Van~Gool.
\newblock Hrda: Context-aware high-resolution domain-adaptive semantic segmentation.
\newblock In \emph{European conference on computer vision}, pages 372--391. Springer, 2022{\natexlab{b}}.

\bibitem[Hoyer et~al.(2023{\natexlab{a}})Hoyer, Dai, Wang, and Van~Gool]{hoyer2023mic}
Lukas Hoyer, Dengxin Dai, Haoran Wang, and Luc Van~Gool.
\newblock Mic: Masked image consistency for context-enhanced domain adaptation.
\newblock In \emph{Proceedings of the IEEE/CVF conference on computer vision and pattern recognition}, pages 11721--11732, 2023{\natexlab{a}}.

\bibitem[Hoyer et~al.(2023{\natexlab{b}})Hoyer, Dai, Wang, Chen, and Van~Gool]{hoyer2023improving}
Lukas Hoyer, Dengxin Dai, Qin Wang, Yuhua Chen, and Luc Van~Gool.
\newblock Improving semi-supervised and domain-adaptive semantic segmentation with self-supervised depth estimation.
\newblock \emph{International Journal of Computer Vision}, 131\penalty0 (8):\penalty0 2070--2096, 2023{\natexlab{b}}.

\bibitem[Hoyer et~al.(2025)Hoyer, Tan, Naeem, Van~Gool, and Tombari]{hoyer2025semivl}
Lukas Hoyer, David~Joseph Tan, Muhammad~Ferjad Naeem, Luc Van~Gool, and Federico Tombari.
\newblock Semivl: Semi-supervised semantic segmentation with vision-language guidance.
\newblock In \emph{European Conference on Computer Vision}, pages 257--275. Springer, 2025.

\bibitem[Ilharco et~al.(2021)Ilharco, Wortsman, Wightman, Gordon, Carlini, Taori, Dave, Shankar, Namkoong, Miller, Hajishirzi, Farhadi, and Schmidt]{ilharco}
Gabriel Ilharco, Mitchell Wortsman, Ross Wightman, Cade Gordon, Nicholas Carlini, Rohan Taori, Achal Dave, Vaishaal Shankar, Hongseok Namkoong, John Miller, Hannaneh Hajishirzi, Ali Farhadi, and Ludwig Schmidt.
\newblock Openclip, 2021.

\bibitem[Kalluri et~al.(2024)Kalluri, Majumder, and Chandraker]{kalluri2024tell}
Tarun Kalluri, Bodhisattwa~Prasad Majumder, and Manmohan Chandraker.
\newblock Tell, don't show!: Language guidance eases transfer across domains in images and videos.
\newblock \emph{arXiv preprint arXiv:2403.05535}, 2024.

\bibitem[Kang et~al.(2020)Kang, Wei, Yang, Zhuang, and Hauptmann]{kang2020pixel}
Guoliang Kang, Yunchao Wei, Yi Yang, Yueting Zhuang, and Alexander Hauptmann.
\newblock Pixel-level cycle association: A new perspective for domain adaptive semantic segmentation.
\newblock \emph{Advances in neural information processing systems}, 33:\penalty0 3569--3580, 2020.

\bibitem[Kim et~al.(2024)Kim, Lee, and Lee]{kim2024lc}
Young-Eun Kim, Yu-Won Lee, and Seong-Whan Lee.
\newblock Lc-msm: Language-conditioned masked segmentation model for unsupervised domain adaptation.
\newblock \emph{Pattern Recognition}, 148:\penalty0 110201, 2024.

\bibitem[Krizhevsky et~al.(2010)Krizhevsky, Nair, and Hinton]{krizhevsky2010cifar}
Alex Krizhevsky, Vinod Nair, and Geoffrey Hinton.
\newblock Cifar-10 (canadian institute for advanced research).
\newblock \emph{URL http://www. cs. toronto. edu/kriz/cifar. html}, 5\penalty0 (4):\penalty0 1, 2010.

\bibitem[Landman et~al.(2015)Landman, Xu, Igelsias, Styner, Langerak, and Klein]{landman2015miccai}
Bennett Landman, Zhoubing Xu, J Igelsias, Martin Styner, Thomas Langerak, and Arno Klein.
\newblock Miccai multi-atlas labeling beyond the cranial vault--workshop and challenge.
\newblock In \emph{Proc. MICCAI Multi-Atlas Labeling Beyond Cranial Vault—Workshop Challenge}, page~12, 2015.

\bibitem[Li et~al.(2022{\natexlab{a}})Li, Weinberger, Belongie, Koltun, and Ranftl]{li2022language}
Boyi Li, Kilian~Q Weinberger, Serge Belongie, Vladlen Koltun, and Ren{\'e} Ranftl.
\newblock Language-driven semantic segmentation.
\newblock \emph{arXiv preprint arXiv:2201.03546}, 2022{\natexlab{a}}.

\bibitem[Li et~al.(2023{\natexlab{a}})Li, Weinberger, Belongie, Koltun, and Ranftl]{lilanguage}
Boyi Li, Kilian~Q Weinberger, Serge Belongie, Vladlen Koltun, and Rene Ranftl.
\newblock Language-driven semantic segmentation.
\newblock In \emph{International Conference on Learning Representations}, 2023{\natexlab{a}}.

\bibitem[Li et~al.(2020)Li, Chen, Ding, Zhu, Lu, and Shen]{li2020maximum}
Jingjing Li, Erpeng Chen, Zhengming Ding, Lei Zhu, Ke Lu, and Heng~Tao Shen.
\newblock Maximum density divergence for domain adaptation.
\newblock \emph{IEEE transactions on pattern analysis and machine intelligence}, 43\penalty0 (11):\penalty0 3918--3930, 2020.

\bibitem[Li et~al.(2021)Li, Li, Shi, and Yu]{li2021cross}
Jichang Li, Guanbin Li, Yemin Shi, and Yizhou Yu.
\newblock Cross-domain adaptive clustering for semi-supervised domain adaptation.
\newblock In \emph{Proceedings of the IEEE/CVF conference on computer vision and pattern recognition}, pages 2505--2514, 2021.

\bibitem[Li et~al.(2023{\natexlab{b}})Li, Li, Savarese, and Hoi]{li2023blip}
Junnan Li, Dongxu Li, Silvio Savarese, and Steven Hoi.
\newblock Blip-2: Bootstrapping language-image pre-training with frozen image encoders and large language models.
\newblock In \emph{International conference on machine learning}, pages 19730--19742. PMLR, 2023{\natexlab{b}}.

\bibitem[Li et~al.(2022{\natexlab{b}})Li, Li, He, Zhang, Jia, and Zhang]{li2022class}
Ruihuang Li, Shuai Li, Chenhang He, Yabin Zhang, Xu Jia, and Lei Zhang.
\newblock Class-balanced pixel-level self-labeling for domain adaptive semantic segmentation.
\newblock In \emph{Proceedings of the IEEE/CVF conference on computer vision and pattern recognition}, pages 11593--11603, 2022{\natexlab{b}}.

\bibitem[Li et~al.(2023{\natexlab{c}})Li, Roy, Zhou, Lu, and Lathuili{\`e}re]{li2023contrast}
Tianyu Li, Subhankar Roy, Huayi Zhou, Hongtao Lu, and St{\'e}phane Lathuili{\`e}re.
\newblock Contrast, stylize and adapt: Unsupervised contrastive learning framework for domain adaptive semantic segmentation.
\newblock In \emph{Proceedings of the IEEE/CVF Conference on Computer Vision and Pattern Recognition}, pages 4869--4879, 2023{\natexlab{c}}.

\bibitem[Li et~al.(2019)Li, Yu, Chang, Ma, and Cao]{li2019dual}
Xiaoxu Li, Liyun Yu, Dongliang Chang, Zhanyu Ma, and Jie Cao.
\newblock Dual cross-entropy loss for small-sample fine-grained vehicle classification.
\newblock \emph{IEEE Transactions on Vehicular Technology}, 68\penalty0 (5):\penalty0 4204--4212, 2019.

\bibitem[Liang et~al.(2023)Liang, Wu, Dai, Li, Zhao, Zhang, Zhang, Vajda, and Marculescu]{liang2023open}
Feng Liang, Bichen Wu, Xiaoliang Dai, Kunpeng Li, Yinan Zhao, Hang Zhang, Peizhao Zhang, Peter Vajda, and Diana Marculescu.
\newblock Open-vocabulary semantic segmentation with mask-adapted clip.
\newblock In \emph{Proceedings of the IEEE/CVF Conference on Computer Vision and Pattern Recognition}, pages 7061--7070, 2023.

\bibitem[Lin(2017)]{lin2017focal}
T Lin.
\newblock Focal loss for dense object detection.
\newblock \emph{arXiv preprint arXiv:1708.02002}, 2017.

\bibitem[Lin et~al.(2022)Lin, Yao, Li, Zheng, and Li]{lin2022calibrating}
Yiqun Lin, Huifeng Yao, Zezhong Li, Guoyan Zheng, and Xiaomeng Li.
\newblock Calibrating label distribution for class-imbalanced barely-supervised knee segmentation.
\newblock In \emph{International Conference on Medical Image Computing and Computer-Assisted Intervention}, pages 109--118. Springer, 2022.

\bibitem[Liu et~al.(2023)Liu, Ding, Zhang, and Jiang]{liu2023multi}
Chang Liu, Henghui Ding, Yulun Zhang, and Xudong Jiang.
\newblock Multi-modal mutual attention and iterative interaction for referring image segmentation.
\newblock \emph{IEEE Transactions on Image Processing}, 32:\penalty0 3054--3065, 2023.

\bibitem[L\"uddecke and Ecker(2022)]{lueddecke22_cvpr}
Timo L\"uddecke and Alexander Ecker.
\newblock Image segmentation using text and image prompts.
\newblock In \emph{Proceedings of the IEEE/CVF Conference on Computer Vision and Pattern Recognition (CVPR)}, pages 7086--7096, 2022.

\bibitem[Luo et~al.(2022)Luo, Wang, Liao, Chen, Song, Chen, Zhang, Metaxas, and Zhang]{luo2022semi}
Xiangde Luo, Guotai Wang, Wenjun Liao, Jieneng Chen, Tao Song, Yinan Chen, Shichuan Zhang, Dimitris~N Metaxas, and Shaoting Zhang.
\newblock Semi-supervised medical image segmentation via uncertainty rectified pyramid consistency.
\newblock \emph{Medical Image Analysis}, 80:\penalty0 102517, 2022.

\bibitem[Ma et~al.(2023)Ma, Wang, Liu, Lin, and Li]{ma2023enhanced}
Jie Ma, Chuan Wang, Yang Liu, Liang Lin, and Guanbin Li.
\newblock Enhanced soft label for semi-supervised semantic segmentation.
\newblock In \emph{Proceedings of the IEEE/CVF International Conference on Computer Vision}, pages 1185--1195, 2023.

\bibitem[Mansour et~al.(2024)Mansour, Unal, Saha, Bejar, and Van~Gool]{mansour2024language}
Elham~Amin Mansour, Ozan Unal, Suman Saha, Benjamin Bejar, and Luc Van~Gool.
\newblock Language-guided instance-aware domain-adaptive panoptic segmentation.
\newblock \emph{arXiv preprint arXiv:2404.03799}, 2024.

\bibitem[Mei et~al.(2020)Mei, Zhu, Zou, and Zhang]{mei2020instance}
Ke Mei, Chuang Zhu, Jiaqi Zou, and Shanghang Zhang.
\newblock Instance adaptive self-training for unsupervised domain adaptation.
\newblock In \emph{Computer Vision--ECCV 2020: 16th European Conference, Glasgow, UK, August 23--28, 2020, Proceedings, Part XXVI 16}, pages 415--430. Springer, 2020.

\bibitem[Milletari et~al.(2016)Milletari, Navab, and Ahmadi]{milletari2016v}
Fausto Milletari, Nassir Navab, and Seyed-Ahmad Ahmadi.
\newblock V-net: Fully convolutional neural networks for volumetric medical image segmentation.
\newblock In \emph{2016 fourth international conference on 3D vision (3DV)}, pages 565--571. Ieee, 2016.

\bibitem[Muhammad et~al.(2022)Muhammad, Hussain, Ullah, Del~Ser, Rezaei, Kumar, Hijji, Bellavista, and de~Albuquerque]{muhammad2022vision}
Khan Muhammad, Tanveer Hussain, Hayat Ullah, Javier Del~Ser, Mahdi Rezaei, Neeraj Kumar, Mohammad Hijji, Paolo Bellavista, and Victor Hugo~C de Albuquerque.
\newblock Vision-based semantic segmentation in scene understanding for autonomous driving: Recent achievements, challenges, and outlooks.
\newblock \emph{IEEE Transactions on Intelligent Transportation Systems}, 23\penalty0 (12):\penalty0 22694--22715, 2022.

\bibitem[Olsson et~al.(2021)Olsson, Tranheden, Pinto, and Svensson]{olsson2021classmix}
Viktor Olsson, Wilhelm Tranheden, Juliano Pinto, and Lennart Svensson.
\newblock Classmix: Segmentation-based data augmentation for semi-supervised learning.
\newblock In \emph{Proceedings of the IEEE/CVF winter conference on applications of computer vision}, pages 1369--1378, 2021.

\bibitem[Qin et~al.(2022)Qin, Wang, and Fu]{qin2022semi}
Can Qin, Yizhou Wang, and Yun Fu.
\newblock Robust semi-supervised domain adaptation against noisy labels.
\newblock In \emph{Proceedings of the 31st ACM International Conference on Information \& Knowledge Management}, pages 4409--4413, 2022.

\bibitem[Qiu et~al.(2023)Qiu, Cheng, Lu, Zhang, Wan, Xue, and Pu]{qiu2023subclassified}
Shoumeng Qiu, Xianhui Cheng, Hong Lu, Haiqiang Zhang, Ru Wan, Xiangyang Xue, and Jian Pu.
\newblock Subclassified loss: Rethinking data imbalance from subclass perspective for semantic segmentation.
\newblock \emph{IEEE Transactions on Intelligent Vehicles}, 2023.

\bibitem[Radford et~al.(2021)Radford, Kim, Hallacy, Ramesh, Goh, Agarwal, Sastry, Askell, Mishkin, Clark, et~al.]{radford2021learning}
Alec Radford, Jong~Wook Kim, Chris Hallacy, Aditya Ramesh, Gabriel Goh, Sandhini Agarwal, Girish Sastry, Amanda Askell, Pamela Mishkin, Jack Clark, et~al.
\newblock Learning transferable visual models from natural language supervision.
\newblock In \emph{International conference on machine learning}, pages 8748--8763. PMLR, 2021.

\bibitem[Ren et~al.(2023)Ren, Tang, Sun, Zhao, and Han]{ren2023visual}
Wenqi Ren, Yang Tang, Qiyu Sun, Chaoqiang Zhao, and Qing-Long Han.
\newblock Visual semantic segmentation based on few/zero-shot learning: An overview.
\newblock \emph{IEEE/CAA Journal of Automatica Sinica}, 2023.

\bibitem[Richter et~al.(2016)Richter, Vineet, Roth, and Koltun]{richter2016playing}
Stephan~R Richter, Vibhav Vineet, Stefan Roth, and Vladlen Koltun.
\newblock Playing for data: Ground truth from computer games.
\newblock In \emph{Computer Vision--ECCV 2016: 14th European Conference, Amsterdam, The Netherlands, October 11-14, 2016, Proceedings, Part II 14}, pages 102--118. Springer, 2016.

\bibitem[Ros et~al.(2016)Ros, Sellart, Materzynska, Vazquez, and Lopez]{ros2016synthia}
German Ros, Laura Sellart, Joanna Materzynska, David Vazquez, and Antonio~M Lopez.
\newblock The synthia dataset: A large collection of synthetic images for semantic segmentation of urban scenes.
\newblock In \emph{Proceedings of the IEEE conference on computer vision and pattern recognition}, pages 3234--3243, 2016.

\bibitem[Saini and Susan(2023)]{saini2023tackling}
Manisha Saini and Seba Susan.
\newblock Tackling class imbalance in computer vision: a contemporary review.
\newblock \emph{Artificial Intelligence Review}, 56\penalty0 (Suppl 1):\penalty0 1279--1335, 2023.

\bibitem[Saito et~al.(2019)Saito, Kim, Sclaroff, Darrell, and Saenko]{saito2019semi}
Kuniaki Saito, Donghyun Kim, Stan Sclaroff, Trevor Darrell, and Kate Saenko.
\newblock Semi-supervised domain adaptation via minimax entropy.
\newblock In \emph{Proceedings of the IEEE/CVF international conference on computer vision}, pages 8050--8058, 2019.

\bibitem[Schwonberg et~al.(2023)Schwonberg, Niemeijer, Term{\"o}hlen, Schmidt, Gottschalk, Fingscheidt, et~al.]{schwonberg2023survey}
Manuel Schwonberg, Joshua Niemeijer, Jan-Aike Term{\"o}hlen, Nico~M Schmidt, Hanno Gottschalk, Tim Fingscheidt, et~al.
\newblock Survey on unsupervised domain adaptation for semantic segmentation for visual perception in automated driving.
\newblock \emph{IEEE Access}, 11:\penalty0 54296--54336, 2023.

\bibitem[Shen et~al.(2023{\natexlab{a}})Shen, Gurram, Liu, Wang, and Knoll]{shen2023diga}
Fengyi Shen, Akhil Gurram, Ziyuan Liu, He Wang, and Alois Knoll.
\newblock Diga: Distil to generalize and then adapt for domain adaptive semantic segmentation.
\newblock In \emph{Proceedings of the IEEE/CVF Conference on Computer Vision and Pattern Recognition}, pages 15866--15877, 2023{\natexlab{a}}.

\bibitem[Shen et~al.(2023{\natexlab{b}})Shen, Peng, Wang, Wang, Cen, Jiang, Xie, Yang, and Tian]{shen2023survey}
Wei Shen, Zelin Peng, Xuehui Wang, Huayu Wang, Jiazhong Cen, Dongsheng Jiang, Lingxi Xie, Xiaokang Yang, and Qi Tian.
\newblock A survey on label-efficient deep image segmentation: Bridging the gap between weak supervision and dense prediction.
\newblock \emph{IEEE transactions on pattern analysis and machine intelligence}, 45\penalty0 (8):\penalty0 9284--9305, 2023{\natexlab{b}}.

\bibitem[Sudre et~al.(2017)Sudre, Li, Vercauteren, Ourselin, and Jorge~Cardoso]{sudre2017generalised}
Carole~H Sudre, Wenqi Li, Tom Vercauteren, Sebastien Ourselin, and M Jorge~Cardoso.
\newblock Generalised dice overlap as a deep learning loss function for highly unbalanced segmentations.
\newblock In \emph{Deep Learning in Medical Image Analysis and Multimodal Learning for Clinical Decision Support: Third International Workshop, DLMIA 2017, and 7th International Workshop, ML-CDS 2017, Held in Conjunction with MICCAI 2017, Qu{\'e}bec City, QC, Canada, September 14, Proceedings 3}, pages 240--248. Springer, 2017.

\bibitem[Tarvainen and Valpola(2017)]{tarvainen2017mean}
Antti Tarvainen and Harri Valpola.
\newblock Mean teachers are better role models: Weight-averaged consistency targets improve semi-supervised deep learning results.
\newblock \emph{Advances in neural information processing systems}, 30, 2017.

\bibitem[Thisanke et~al.(2023)Thisanke, Deshan, Chamith, Seneviratne, Vidanaarachchi, and Herath]{thisanke2023semantic}
Hans Thisanke, Chamli Deshan, Kavindu Chamith, Sachith Seneviratne, Rajith Vidanaarachchi, and Damayanthi Herath.
\newblock Semantic segmentation using vision transformers: A survey.
\newblock \emph{Engineering Applications of Artificial Intelligence}, 126:\penalty0 106669, 2023.

\bibitem[Tranheden et~al.(2021)Tranheden, Olsson, Pinto, and Svensson]{tranheden2021dacs}
Wilhelm Tranheden, Viktor Olsson, Juliano Pinto, and Lennart Svensson.
\newblock Dacs: Domain adaptation via cross-domain mixed sampling.
\newblock In \emph{Proceedings of the IEEE/CVF winter conference on applications of computer vision}, pages 1379--1389, 2021.

\bibitem[Truong et~al.(2023)Truong, Le, Raj, Cothren, and Luu]{truong2023fredom}
Thanh-Dat Truong, Ngan Le, Bhiksha Raj, Jackson Cothren, and Khoa Luu.
\newblock Fredom: Fairness domain adaptation approach to semantic scene understanding.
\newblock In \emph{Proceedings of the IEEE/CVF Conference on Computer Vision and Pattern Recognition}, pages 19988--19997, 2023.

\bibitem[Vaswani(2017)]{vaswani2017attention}
A Vaswani.
\newblock Attention is all you need.
\newblock \emph{Advances in Neural Information Processing Systems}, 2017.

\bibitem[Wang and Li(2023)]{wang2023dhc}
Haonan Wang and Xiaomeng Li.
\newblock Dhc: Dual-debiased heterogeneous co-training framework for class-imbalanced semi-supervised medical image segmentation.
\newblock In \emph{International Conference on Medical Image Computing and Computer-Assisted Intervention}, pages 582--591. Springer, 2023.

\bibitem[Wang and Li(2024)]{wang2024towards}
Haonan Wang and Xiaomeng Li.
\newblock Towards generic semi-supervised framework for volumetric medical image segmentation.
\newblock \emph{Advances in Neural Information Processing Systems}, 36, 2024.

\bibitem[Wang et~al.(2020)Wang, Shen, Zhang, Duan, and Mei]{wang2020classes}
Haoran Wang, Tong Shen, Wei Zhang, Ling-Yu Duan, and Tao Mei.
\newblock Classes matter: A fine-grained adversarial approach to cross-domain semantic segmentation.
\newblock In \emph{European conference on computer vision}, pages 642--659. Springer, 2020.

\bibitem[Wang et~al.(2022{\natexlab{a}})Wang, Wu, Lian, and Yu]{wang2022debiased}
Xudong Wang, Zhirong Wu, Long Lian, and Stella~X Yu.
\newblock Debiased learning from naturally imbalanced pseudo-labels.
\newblock In \emph{Proceedings of the IEEE/CVF Conference on Computer Vision and Pattern Recognition}, pages 14647--14657, 2022{\natexlab{a}}.

\bibitem[Wang et~al.(2022{\natexlab{b}})Wang, Wang, Shen, Fei, Li, Jin, Wu, Zhao, and Le]{wang2022semiu2pl}
Yuchao Wang, Haochen Wang, Yujun Shen, Jingjing Fei, Wei Li, Guoqiang Jin, Liwei Wu, Rui Zhao, and Xinyi Le.
\newblock Semi-supervised semantic segmentation using unreliable pseudo-labels.
\newblock In \emph{Proceedings of the IEEE/CVF conference on computer vision and pattern recognition}, pages 4248--4257, 2022{\natexlab{b}}.

\bibitem[Wei et~al.(2021)Wei, Sohn, Mellina, Yuille, and Yang]{wei2021crest}
Chen Wei, Kihyuk Sohn, Clayton Mellina, Alan Yuille, and Fan Yang.
\newblock Crest: A class-rebalancing self-training framework for imbalanced semi-supervised learning.
\newblock In \emph{Proceedings of the IEEE/CVF conference on computer vision and pattern recognition}, pages 10857--10866, 2021.

\bibitem[Wen et~al.(2024)Wen, Xu, Feng, Zhou, Zhou, and Wang]{wen2024semi}
Lu Wen, Yuanyuan Xu, Zhenghao Feng, Jiliu Zhou, Luping Zhou, and Yan Wang.
\newblock Semi-supervised domain adaptation for semantic segmentation via active learning with feature-and semantic-level alignments.
\newblock \emph{IEEE Transactions on Intelligent Vehicles}, 2024.

\bibitem[Wu and Chan(2024)]{wu98gradient}
Wenbo Qi1~Jiafei Wu and SC Chan.
\newblock Gradient-aware for class-imbalanced semi-supervised medical image segmentation.
\newblock \emph{ECCV}, 98:\penalty0 1--86, 2024.

\bibitem[Wu et~al.(2022)Wu, Wu, Wu, Ge, and Cai]{wu2022exploring}
Yicheng Wu, Zhonghua Wu, Qianyi Wu, Zongyuan Ge, and Jianfei Cai.
\newblock Exploring smoothness and class-separation for semi-supervised medical image segmentation.
\newblock In \emph{International conference on medical image computing and computer-assisted intervention}, pages 34--43. Springer, 2022.

\bibitem[Wu et~al.(2024)Wu, Xing, Zhang, Xie, and Qu]{wu2024clip2uda}
Yao Wu, Mingwei Xing, Yachao Zhang, Yuan Xie, and Yanyun Qu.
\newblock Clip2uda: Making frozen clip reward unsupervised domain adaptation in 3d semantic segmentation.
\newblock In \emph{Proceedings of the 32nd ACM International Conference on Multimedia}, pages 8662--8671, 2024.

\bibitem[Wu et~al.(2016)Wu, Shen, and Hengel]{wu2016bridging}
Zifeng Wu, Chunhua Shen, and Anton van~den Hengel.
\newblock Bridging category-level and instance-level semantic image segmentation.
\newblock \emph{arXiv preprint arXiv:1605.06885}, 2016.

\bibitem[Xie et~al.(2022)Xie, Yuan, Li, Liu, and Cheng]{xie2022towards}
Binhui Xie, Longhui Yuan, Shuang Li, Chi~Harold Liu, and Xinjing Cheng.
\newblock Towards fewer annotations: Active learning via region impurity and prediction uncertainty for domain adaptive semantic segmentation.
\newblock In \emph{Proceedings of the IEEE/CVF conference on computer vision and pattern recognition}, pages 8068--8078, 2022.

\bibitem[Xu et~al.(2022)Xu, De~Mello, Liu, Byeon, Breuel, Kautz, and Wang]{xu2022groupvit}
Jiarui Xu, Shalini De~Mello, Sifei Liu, Wonmin Byeon, Thomas Breuel, Jan Kautz, and Xiaolong Wang.
\newblock Groupvit: Semantic segmentation emerges from text supervision.
\newblock In \emph{Proceedings of the IEEE/CVF Conference on Computer Vision and Pattern Recognition}, pages 18134--18144, 2022.

\bibitem[Yan et~al.(2019)Yan, Li, Wang, Li, Xu, and Zuo]{yan2019weighted}
Hongliang Yan, Zhetao Li, Qilong Wang, Peihua Li, Yong Xu, and Wangmeng Zuo.
\newblock Weighted and class-specific maximum mean discrepancy for unsupervised domain adaptation.
\newblock \emph{IEEE Transactions on Multimedia}, 22\penalty0 (9):\penalty0 2420--2433, 2019.

\bibitem[Yan et~al.(2024)Yan, Qian, Li, Li, Wang, and Yang]{yan2024ss}
Weihao Yan, Yeqiang Qian, Yueyuan Li, Tao Li, Chunxiang Wang, and Ming Yang.
\newblock Ss-ada: A semi-supervised active domain adaptation framework for semantic segmentation.
\newblock \emph{arXiv preprint arXiv:2407.12788}, 2024.

\bibitem[Yang et~al.(2021{\natexlab{a}})Yang, An, Yan, Zhao, and Huang]{yang2021context}
Jinyu Yang, Weizhi An, Chaochao Yan, Peilin Zhao, and Junzhou Huang.
\newblock Context-aware domain adaptation in semantic segmentation.
\newblock In \emph{Proceedings of the IEEE/CVF Winter Conference on Applications of Computer Vision}, pages 514--524, 2021{\natexlab{a}}.

\bibitem[Yang et~al.(2021{\natexlab{b}})Yang, Wang, Gao, Shrivastava, Weinberger, Chao, and Lim]{yang2021deep}
Luyu Yang, Yan Wang, Mingfei Gao, Abhinav Shrivastava, Kilian~Q Weinberger, Wei-Lun Chao, and Ser-Nam Lim.
\newblock Deep co-training with task decomposition for semi-supervised domain adaptation.
\newblock In \emph{Proceedings of the IEEE/CVF international conference on computer vision}, pages 8906--8916, 2021{\natexlab{b}}.

\bibitem[Yang et~al.(2023)Yang, Qi, Feng, Zhang, and Shi]{yang2023revisiting}
Lihe Yang, Lei Qi, Litong Feng, Wayne Zhang, and Yinghuan Shi.
\newblock Revisiting weak-to-strong consistency in semi-supervised semantic segmentation.
\newblock In \emph{Proceedings of the IEEE/CVF Conference on Computer Vision and Pattern Recognition}, pages 7236--7246, 2023.

\bibitem[Yu et~al.(2020)Yu, Wang, Li, Fu, and Heng]{yu2019uncertainty}
Lequan Yu, Shujun Wang, Xiaomeng Li, Chi-Wing Fu, and Pheng-Ann Heng.
\newblock Uncertainty-aware self-ensembling model for semi-supervised 3d left atrium segmentation.
\newblock In \emph{Medical image computing and computer assisted intervention--MICCAI 2019: 22nd international conference, Shenzhen, China, October 13--17, 2019, proceedings, part II 22}, pages 605--613. Springer, 2020.

\bibitem[Yu et~al.(2023)Yu, Yang, Huang, Wang, and Yang]{yu2023high}
Lei Yu, Wanqi Yang, Shengqi Huang, Lei Wang, and Ming Yang.
\newblock High-level semantic feature matters few-shot unsupervised domain adaptation.
\newblock In \emph{Proceedings of the AAAI Conference on Artificial Intelligence}, pages 11025--11033, 2023.

\bibitem[Yu and Lin(2023)]{yu2023semi}
Yu-Chu Yu and Hsuan-Tien Lin.
\newblock Semi-supervised domain adaptation with source label adaptation.
\newblock In \emph{Proceedings of the IEEE/CVF Conference on Computer Vision and Pattern Recognition}, pages 24100--24109, 2023.

\bibitem[Zhang et~al.(2023)Zhang, Xie, Ding, and Wang]{zhang2023cross}
Jing Zhang, Yingshuai Xie, Weichao Ding, and Zhe Wang.
\newblock Cross on cross attention: Deep fusion transformer for image captioning.
\newblock \emph{IEEE Transactions on Circuits and Systems for Video Technology}, 33\penalty0 (8):\penalty0 4257--4268, 2023.

\bibitem[Zhang et~al.(2021)Zhang, Zhang, Zhang, Chen, Wang, and Wen]{zhang2021prototypical}
Pan Zhang, Bo Zhang, Ting Zhang, Dong Chen, Yong Wang, and Fang Wen.
\newblock Prototypical pseudo label denoising and target structure learning for domain adaptive semantic segmentation.
\newblock In \emph{Proceedings of the IEEE/CVF conference on computer vision and pattern recognition}, pages 12414--12424, 2021.

\bibitem[Zhang and Sabuncu(2018)]{zhang2018generalized}
Zhilu Zhang and Mert Sabuncu.
\newblock Generalized cross entropy loss for training deep neural networks with noisy labels.
\newblock \emph{Advances in neural information processing systems}, 31, 2018.

\bibitem[Zhou et~al.(2022)Zhou, Loy, and Dai]{zhou2022extract}
Chong Zhou, Chen~Change Loy, and Bo Dai.
\newblock Extract free dense labels from clip.
\newblock In \emph{European Conference on Computer Vision}, pages 696--712. Springer, 2022.

\bibitem[Zhou et~al.(2023{\natexlab{a}})Zhou, Dong, Gan, Wang, and Wei]{zhou2023non}
Jinghao Zhou, Li Dong, Zhe Gan, Lijuan Wang, and Furu Wei.
\newblock Non-contrastive learning meets language-image pre-training.
\newblock In \emph{Proceedings of the IEEE/CVF Conference on Computer Vision and Pattern Recognition}, pages 11028--11038, 2023{\natexlab{a}}.

\bibitem[Zhou et~al.(2023{\natexlab{b}})Zhou, Lei, Zhang, Liu, and Liu]{zhou2023zegclip}
Ziqin Zhou, Yinjie Lei, Bowen Zhang, Lingqiao Liu, and Yifan Liu.
\newblock Zegclip: Towards adapting clip for zero-shot semantic segmentation.
\newblock In \emph{Proceedings of the IEEE/CVF Conference on Computer Vision and Pattern Recognition}, pages 11175--11185, 2023{\natexlab{b}}.

\bibitem[Zhou et~al.(2023{\natexlab{c}})Zhou, Zheng, Liu, Tian, Chen, Chen, and Dong]{zhou2023dynamic}
Zheng Zhou, Change Zheng, Xiaodong Liu, Ye Tian, Xiaoyi Chen, Xuexue Chen, and Zixun Dong.
\newblock A dynamic effective class balanced approach for remote sensing imagery semantic segmentation of imbalanced data.
\newblock \emph{Remote Sensing}, 15\penalty0 (7):\penalty0 1768, 2023{\natexlab{c}}.

\bibitem[Zou et~al.(2018)Zou, Yu, Kumar, and Wang]{zou2018unsupervised}
Yang Zou, Zhiding Yu, BVK Kumar, and Jinsong Wang.
\newblock Unsupervised domain adaptation for semantic segmentation via class-balanced self-training.
\newblock In \emph{Proceedings of the European conference on computer vision (ECCV)}, pages 289--305, 2018.

\bibitem[Zou et~al.(2023)Zou, Chen, Shi, Guo, and Ye]{zou2023object}
Zhengxia Zou, Keyan Chen, Zhenwei Shi, Yuhong Guo, and Jieping Ye.
\newblock Object detection in 20 years: A survey.
\newblock \emph{Proceedings of the IEEE}, 111\penalty0 (3):\penalty0 257--276, 2023.

\end{thebibliography}
